\PassOptionsToPackage{table}{xcolor}
\PassOptionsToPackage{pagebackref,breaklinks=true,colorlinks,citecolor=blue,urlcolor=blue,linkcolor=blue,bookmarks=false}{hyperref}
\documentclass[onecolumn,numbers]{april_aigc}

\usepackage{amsmath,amsfonts,bm}

\def\eqref#1{equation~\ref{#1}}

\def\1{\bm{1}}

\DeclareMathAlphabet{\mathsfit}{\encodingdefault}{\sfdefault}{m}{sl}
\SetMathAlphabet{\mathsfit}{bold}{\encodingdefault}{\sfdefault}{bx}{n}

\usepackage{amsfonts,amssymb,amsmath,graphicx,multirow,booktabs,colortbl,array,wrapfig,placeins,etoolbox,caption}
\usepackage{url}
\usepackage{makecell}

\titleformat*{\section}{\color{aprilblue}\Large\sffamily\bfseries}
\titleformat*{\subsection}{\color{aprilblue}\large\sffamily\bfseries}
\titleformat*{\subsubsection}{\color{aprilblue}\normalsize\sffamily\bfseries}
\DeclareCaptionFormat{aprilcaption}{{\color{aprilblue}\sffamily\textbf{#1 #2}} #3}
\usepackage{fancyhdr}
\newcommand{\pvaprilheaderlogo}{%
  \IfFileExists{assets/april_logo1.png}{%
    \includegraphics[height=0.68cm]{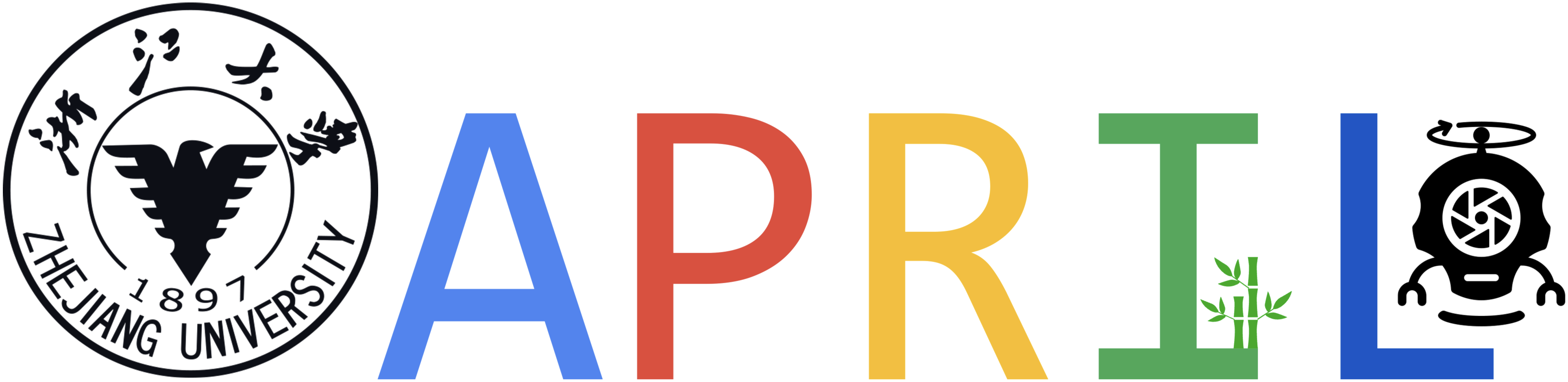}%
  }{}%
}
\renewcommand{\headrulewidth}{0pt}
\renewcommand{\footrulewidth}{0pt}
\definecolor{linkcolor}{named}{aprilblue}
\definecolor{urlcolor}{RGB}{255,105,180}
\definecolor{citecolor}{RGB}{66,168,235}
\hypersetup{colorlinks=true,linkcolor=linkcolor,urlcolor=urlcolor,citecolor=citecolor,bookmarks=false}

\newcommand{\cmark}{\checkmark}
\newcommand{\xmark}{$\times$}
\newcommand{\insightbox}[3]{%
  \par\vspace{0pt}\noindent
  \setlength{\fboxsep}{2pt}%
  \colorbox{black!4}{\parbox{\dimexpr\linewidth-2\fboxsep\relax}{%
    \textcolor{blue!55!black}{\rule{1.5pt}{2.4ex}}\hspace{0.65em}%
    {\color{blue!55!black}\textsc{\bfseries Insight~#1.}}\ \textit{\textbf{#2.} #3}%
  }}%
  \par\vspace{0pt}
}
\newcommand{\bench}{PVBench}
\title{{\bench}: Quantifying Physical Plausibility in Video Generation with Object-Level Physical Measurements}
\author[1]{Kaiwei Chen}
\author[2]{Moran Li}
\author[3]{Zhenye Gan}
\author[3]{Yabiao Wang}
\author[3]{Chengjie Wang}
\author[1]{Kexin Zhang}
\author[1]{Yong Liu}
\author[1]{Jiangning Zhang}
\affiliation[1]{APRIL Lab, Zhejiang University}
\affiliation[2]{Huazhong University of Science and Technology}
\affiliation[3]{Tencent Youtu Lab}
\abstract{Recent video generation models produce visually compelling outputs, yet their physical plausibility remains poorly measured. Physical plausibility asks whether generated objects obey geometry, kinematics, dynamics, and conservation laws. Existing benchmarks rely on coarse visual-language model judgments or target embodied world models rather than video generation itself. We introduce \textbf{{\bench}}, \textbf{\textit{a physical video benchmark that quantitatively measures the plausibility of generated videos across 46 dimensions organized into seven domains}}, spanning: 
\textbf{\textit{i)}} instruction and semantic fidelity, 
\textbf{\textit{ii)}} behavioral and causal consistency, 
\textbf{\textit{iii)}} perceptual and temporal quality, 
\textbf{\textit{iv)}} geometric and spatial consistency, 
\textbf{\textit{v)}} material properties and field states, 
\textbf{\textit{vi)}} kinematic and spatiotemporal persistence, 
\textbf{\textit{vii)}} and mechanical plausibility with conservation laws. \textbf{{\bench}} comprises 600 carefully crafted prompts, of which 30\% involve physical embodied agents such as robot arms, humanoids, quadrupeds, and multi-agent systems. For challenging physically tractable dimensions, it pairs each with specialized vision models that derive verifiable, physically grounded measurements, such as object-level 6D pose, metric scale, surface normals, kinematics, and numerical conservation-law residuals. It thus avoids relying solely on subjective scoring, while supporting both text-to-video and image-to-video generation. 
Finally, we validate the agreement between automatic metrics and human judgments through human evaluation. Meanwhile, we evaluate mainstream open-source and closed-source models and present several insights. By releasing the benchmark, prompts, and calibration protocol, we aim to foster physically grounded video generation.
}
\coverdate{October 19, 2026}
\covercorrespondence{\email{12632067@zju.edu.cn}}
\coversourcecode{https://github.com/KSZG-11/PVBench}
\coverproject{https://kszg-11.github.io/PVBench-Website/}

\begin{document}
\maketitle
\thispagestyle{plain}

\section{Introduction}
\label{sec:intro}

Video generation technology has advanced rapidly in recent years. Large-scale Text-/Image-to-Video (T2V/I2V) models can now produce visually convincing videos and are increasingly regarded as an important foundation for world models~\citep{lecun2022path,teng2025magi1}. Yet looking real does not equal obeying physical laws~\citep{motamed2025physicsiq}: generated videos still frequently exhibit objects clipping through each other, vanishing out of nowhere, or wrong scales. For embodied intelligence~\citep{nvidia2025cosmos}, content creation~\citep{brooks2024video}, and simulation~\citep{bruce2024genie}, such physical errors directly undermine usability. 
Therefore, \textbf{\textit{how to measure the physical plausibility of generated videos is a question that video generation must answer before reaching downstream applications.}}

Existing benchmarks still cannot answer this question well. General-purpose video benchmarks~\citep{huang2024vbench,huang2025vbenchpp} mainly assess semantic and perceptual quality; a video that seriously violates physical laws can still score high on these metrics as long as the picture is clear and aligned with the text. Benchmarks targeting physical plausibility have two limitations: 
\textbf{\textit{1)} Semantic binary judgment}: most works~\citep{bansal2025videophy2,gu2026phyworldbench} rely on MLLM or human yes/no ratings, which cannot localize which physical quantity is violated, which object violates physics, let alone quantify the severity of the violation; \textbf{\textit{2)} Insufficient physical-quantity coverage}: existing evaluation sets usually cover only a few commonsense categories or geometric quantities in simple scenarios, yet some object-level physical quantities are crucial for downstream applications. For example, metric scale determines whether a generated video can support robot manipulation and scene reconstruction that require real dimensions, and 6D pose is the interface quantity from ``watching a video'' to ``using a video'' for motion planning and simulation initialization, which is essentially missing in existing works. Recent works have shown the trend of physical-quantity evaluation: Physics-IQ~\citep{motamed2025physicsiq} indirectly characterizes motion with proxy metrics, WorldScore~\citep{duan2025worldscore} introduces geometric measurement, but neither is yet a systematic physical-quantity evaluation framework.

We argue that \textbf{\textit{physical plausibility should not stay at the level of semantic impression, but should be implemented as computable physical quantities and tested against physical laws}}. To this end, this paper proposes PVBench, a physical-plausibility benchmark for video generation models. Its contributions are threefold: 
\textbf{\textit{1)} Systematic physical taxonomy}: it organizes \textbf{46} evaluation dimensions into \textbf{7} domains, of which \textbf{25} physical dimensions span object-intrinsic properties, relations between objects and space, and object dynamics across space and time, covering most physical dimensions that have appeared in existing benchmarks; 
\textbf{\textit{2)} Same-scene T2V/I2V paired prompt set}: 600 prompt groups, each containing T2V/I2V versions of the same scene, 30\% of which involve embodied-agent scenarios, and each prompt binds its target evaluation dimensions at construction time and requires the key physical quantities to be visible in the frame; 
\textbf{\textit{3)} Physical-quantity-dominated evaluation protocol}: specialized vision models extract numerical physical quantities, which are checked against physical laws and then mapped to evaluation scores, while dimensions that cannot be directly computed (\textit{e.g.}, materials, phase change, \textit{etc.}) are supplemented by MLLM scoring.

Experiments show that the physical dimensions of {\bench} are generally consistent with human judgments. Meanwhile, across 6 T2V and 6 I2V mainstream model configurations, the physical-layer scores are substantially lower overall than those in the semantic and perceptual layers, highlighting a persistent gap in physical plausibility. We will open-source the code and release the prompt set to foster physically grounded video generation research. We hope PVBench can push video generation from ``looking good'' to ``physically usable''.

\begin{figure}[!htbp]
  \centering
  \includegraphics[width=\linewidth,trim={91.1bp 195bp 59.4bp 167.8bp},clip]{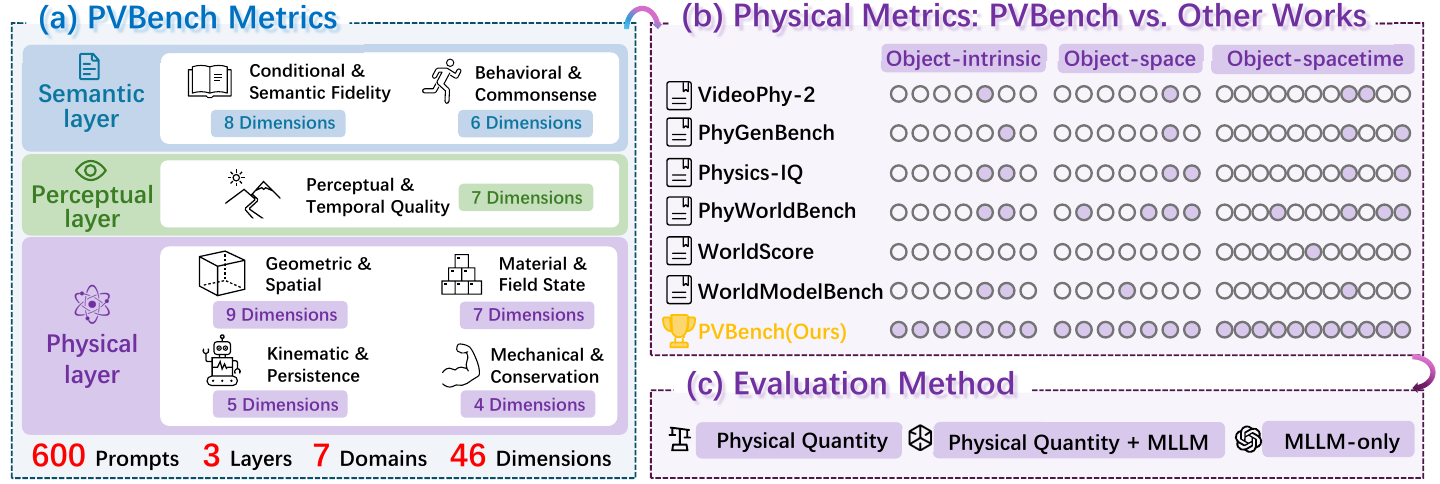}
  \caption{\textbf{Overview of {\bench}.} 
  \textbf{\textit{(a)}} The benchmark comprises 46 evaluation dimensions organized into seven domains and three assessment layers: semantic, perceptual, and physical. 
  \textbf{\textit{(b)}} The physical layer contains 25 dimensions covering object-intrinsic properties, relations between objects and space, and object dynamics across space and time, encompassing most physical dimensions appearing in prior work. 
  \textbf{\textit{(c)}} PVBench uses three evaluation paradigms: physical-quantity measurement, MLLM evaluation, and a combination of both.}
  \label{fig:overview}
\end{figure}\FloatBarrier

\section{Related Work}
\label{sec:related}

\subsection{Video Foundation Models}

Video generation has advanced rapidly under the diffusion paradigm~\citep{ho2020ddpm,song2020score}, with diffusion transformers~\citep{peebles2023dit} and flow matching~\citep{lipman2023flow,liu2023flow} as the dominant architectural and training choices. Systems such as Sora~\citep{brooks2024video}, Wan~\citep{wang2025wan}, and HunyuanVideo 1.5~\citep{wu2025hunyuanvideo15} now produce high-fidelity T2V and image-to-video (I2V) clips at scale, while recent models further diversify the design space. Seedance 2.0~\citep{seed2026seedance2} unifies text, image, audio, and video inputs in a single model, SANA-Video~\citep{chen2025sanavideo} pursues efficient long-video generation with linear attention, and MAGI-1~\citep{teng2025magi1} departs from diffusion altogether to explore autoregressive video generation. 
Beyond media synthesis, video generators are increasingly framed as world models~\citep{lecun2022path}. Cosmos~\citep{nvidia2025cosmos} builds world foundation models as digital twins of the physical world for physical AI, while interactive systems accept action inputs, ranging from latent-action environments~\citep{bruce2024genie} to open-domain action-controlled game generation~\citep{yu2025gamefactory}.

\subsection{Benchmarks for Video Generation}

General-purpose suites decompose video quality into perceptual and semantic metrics computed with pretrained perception models or MLLMs as judges. VBench and VBench++~\citep{huang2024vbench,huang2025vbenchpp} cover temporal consistency, motion smoothness, and text-video alignment, while EvalCrafter~\citep{liu2024evalcrafter} and T2V-CompBench~\citep{sun2024t2vcompbench} aggregate human-preference and compositional-alignment metrics. VBench-2.0~\citep{zheng2025vbench2} moves toward \emph{intrinsic faithfulness} by adding physics, commonsense, and human-fidelity dimensions. Its physics checks remain coarse, however, consisting mainly of VLM question answering on state changes (mechanics, thermotics, material), with feature-matching geometry for multi-view consistency. More recent world-model benchmarks broaden the evaluation scope beyond visual quality. HappyWorld-Bench~\citep{bai2026happyworldbench} evaluates video, spatial, and embodied world models, with an emphasis on consistency and responsiveness during exploration, interaction, and modification.

\textbf{\textit{A parallel line of work targets physical plausibility directly.}} VideoPhy-2~\citep{bansal2025videophy2} and PhyWorldBench~\citep{gu2026phyworldbench} rely on human or MLLM semantic judgments over action-centric activities and diverse prompt categories; PhyGenBench~\citep{meng2025phygenbench} scores 27 physical principles with a hierarchical VLM evaluator; WorldModelBench~\citep{li2025worldmodelbench} rates five physical laws as binary outcomes; Physics-IQ~\citep{motamed2025physicsiq} measures motion-mask overlap on real-scene continuation, with metrics the authors acknowledge remain proxies rather than direct physical quantities. WorldScore~\citep{duan2025worldscore} is the most quantitative, reconstructing camera trajectories with DROID-SLAM~\citep{teed2021droid} and computing reprojection error, dense depth, and optical-flow AEPE. However, these physical-plausibility benchmarks are still limited to semantic judgments and coarse-grained categories, and the few quantitative ones measure only camera-centric geometry. To address these challenges, we evaluate generated videos through numerical physical quantities extracted by specialized models and checked against physical laws.

\section{Methodology}
\label{sec:method}

\subsection{PVBench Evaluation System}
\label{sec:method:overview}

PVBench evaluates generated videos along 46 dimensions. These 46 dimensions are organized into 7 domains and span three assessment layers: semantic, perceptual, and physical. The physical layer encompasses 4 domains with 25 dimensions in total, which are further divided into three families: object-intrinsic physical properties, relations between objects and space, and object dynamics across space and time, covering most common physical quantities. The overall framework is shown in Fig.~\ref{fig:overview}; the complete correspondence is given in Appendix~\ref{sec:appendix:dimcat}, Table~\ref{tab:dimcat}.

PVBench turns ``physical plausibility'' from a qualitative question into a numerically testable one. Except for dimensions that are difficult to quantify, each physical dimension is paired with a computable physical quantity and a physical-law consistency criterion. Semantic and perceptual dimensions follow or adapt existing paradigms, allowing the physical score to be interpreted jointly with semantic fidelity and visual quality.

Compared with prior works, PVBench covers physical dimensions more comprehensively, evaluates primarily through quantitative computation of physical quantities, and uses MLLM to supplement dimensions that are hard to quantify.

\subsection{Prompt Design for Physical Plausibility Evaluation}
\label{sec:method:prompt}

\begin{figure}[!htbp]
  \centering
  \includegraphics[width=\linewidth,trim={42bp 139.0bp 55bp 135bp},clip]{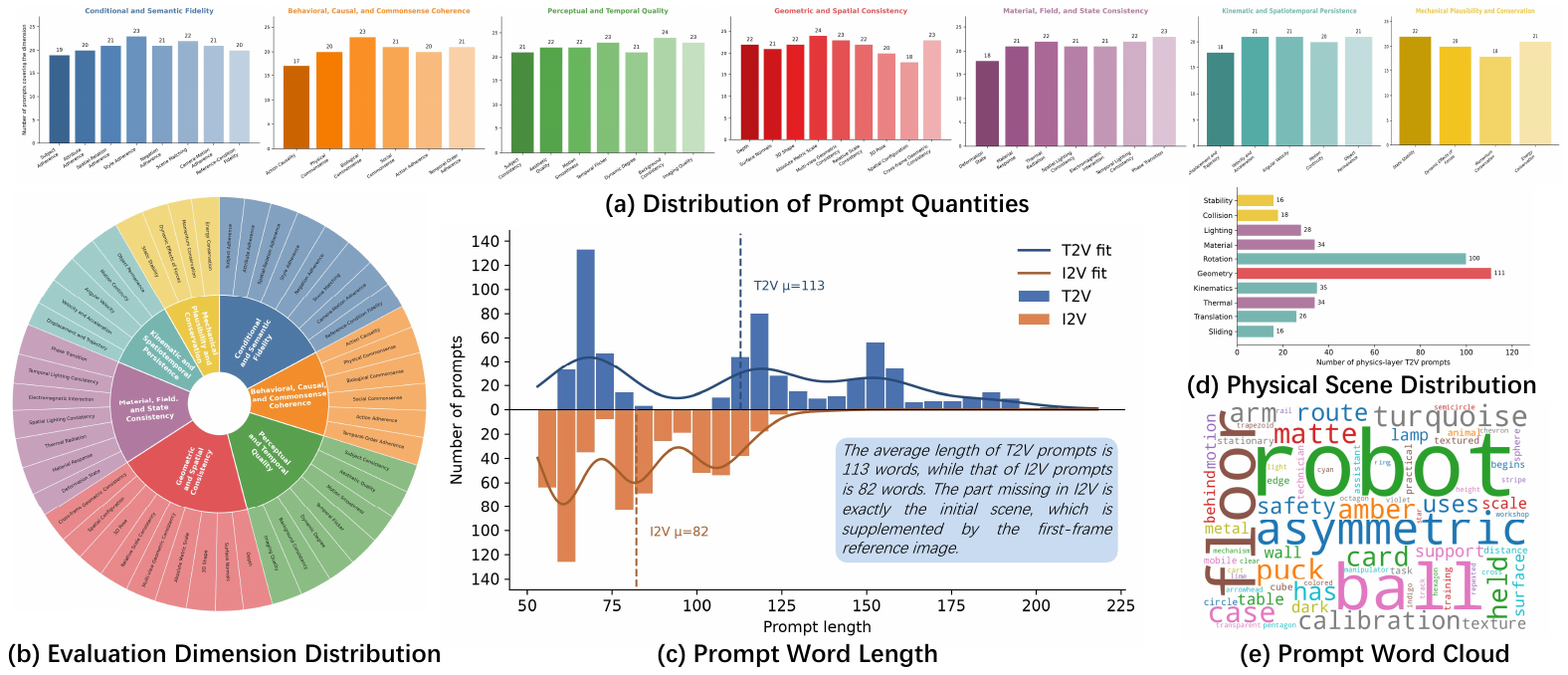}
  \caption{\textbf{Prompt statistics of PVBench.} The figure provides an integrated characterization of the PVBench prompt set, summarizing its dimension distribution, length variation, physical scene coverage, and key vocabulary.}
  \label{fig:distributions}
\end{figure}

\noindent\textbf{Diverse Dedicated Scene Construction.} Prompts fall into three categories: 
\textbf{\textit{1)}} for each of the 46 evaluation dimensions, 5 tailored prompts are designed; 
\textbf{\textit{2)}} related dimensions are paired in all 23 dimension pairs, each pair corresponding to 10 hybrid prompts; 
\textbf{\textit{3)}} embodied agents are injected, together with controlled occlusion, camera motion, and other visual confounders, as a high-difficulty supplement. The three categories total 600 prompt groups, each containing a one-to-one T2V and I2V variant, of which 30\% involve embodied-agent scenarios (Fig.~\ref{fig:distributions}).

\begin{figure}[!htbp]
  \centering
  \includegraphics[width=\linewidth,trim={99.7bp 212.4bp 91.8bp 162.7bp},clip]{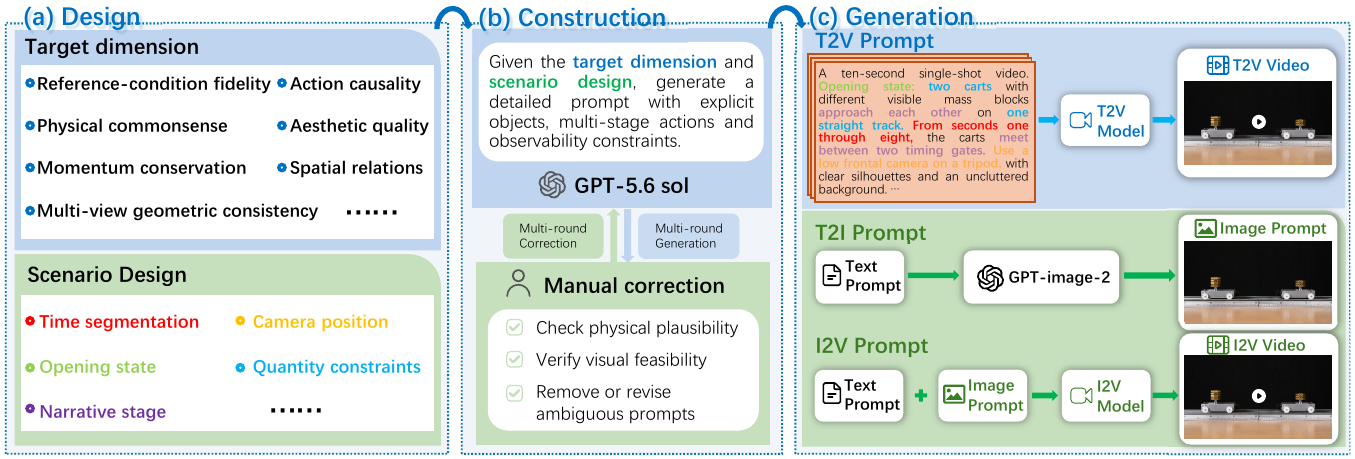}
  \caption{\textbf{PVBench data curation.} \textbf{\textit{(a)}} Design of the target dimensions and scene. \textbf{\textit{(b)}} Construction of prompts through multiple rounds of MLLM assistance and human verification. \textbf{\textit{(c)}} Generation of T2V and I2V videos.}
  \label{fig:prompt_pipeline}
\end{figure}

\noindent\textbf{LLM-Assisted Prompt Construction.} Each prompt group is built in three steps: \textbf{\textit{1)}} given the target dimensions and scene specification, draft the prompt with GPT-5.6 Sol~\citep{openai2026gpt5.6}, explicitly instantiating objects, multi-stage actions, and observability constraints; \textbf{\textit{2)}} each draft is manually screened for physical plausibility, visual feasibility, and unambiguous observability, modified or discarded if necessary; \textbf{\textit{3)}} the reference image of the I2V variant is synthesized by the GPT-image-2 model~\citep{openai2026gptimage2} from the first-frame description of the prompt, so that the T2V and I2V variants share the same initial scene (Fig.~\ref{fig:prompt_pipeline}).

\FloatBarrier

\subsection{Evaluation Protocol}
\label{sec:method:protocol}

\subsubsection{Hierarchical Metrics of PVBench}
\label{sec:method:metrics}

PVBench metrics have three layers: semantic, perceptual, and physical. They are subdivided by evaluation target into 7 domains for a total of 46 dimensions. Table~\ref{tab:modeltodimension} summarizes the dimensions covered by each specialized evaluation model.

\textbf{Semantic layer} examines the consistency between the generated video and the conditioning information and commonsense logic, covering 2 domains and 14 dimensions. Condition and semantic fidelity (CSF) measures whether the video is faithful to the input condition, covering instruction following, attribute correctness, spatial relations, style, negation, scene, camera motion, and reference-condition following. Behavioral, causal, and commonsense consistency (BC) measures whether actions and timing unfold as instructed, and whether behavior conforms to physical, biological, and social commonsense.

The \textbf{perceptual layer}, termed Perceptual and Temporal Quality (PTQ), evaluates the visual quality of video frames and temporal dynamics across 7 dimensions. It covers subject consistency, aesthetic quality, motion smoothness, temporal flickering, dynamic degree, background and imaging quality. The semantic and perceptual layers follow or adapt existing evaluation paradigms, so that the physical-layer score can be interpreted jointly with semantic fidelity and image quality.

\textbf{Physical layer} is the core of PVBench, covering 4 domains and 25 dimensions, divided by the type of physical quantity measured. Geometric and spatial consistency (GS) evaluates the spatial self-consistency of geometric quantities such as depth, normals, shape, scale, pose, spatial relations, and cross-frame geometry. Material properties and field states (MFS) evaluates material attributes and field quantities such as deformation, material, thermal radiation, light/shadow, magnetic/electric, and phase change. Kinematics and spatiotemporal persistence (KP) evaluates kinematic quantities such as displacement, velocity, and angular velocity, as well as spatiotemporal persistence properties such as motion continuity and object permanence. Mechanical plausibility and conservation laws (MC) evaluates mechanical and conserved quantities such as static stability, dynamic effects of forces, momentum conservation, and energy conservation.

\begin{table}[!htbp]
\caption{\textbf{The dimensions covered by each specialized evaluation model.} Dimension names are abbreviated as codes to save space; see the Appendix for the full correspondence.}
\label{tab:modeltodimension}
\vspace{-8pt}
\centering
\begin{minipage}[t]{0.47\textwidth}
\centering
\tiny
\setlength{\tabcolsep}{2pt}
\begin{tabular}{l >{\raggedright\arraybackslash}p{0.7\textwidth}}
\toprule
\multicolumn{1}{c}{\bf MODEL} & \multicolumn{1}{c}{\bf COVERED DIMENSIONS} \\
\hline
VGGT-$\Omega$ & R01, R03, R08, R09, R11, R17, R18, R19 \\
\hline
Depth Anything 3 & R01, R04, R22, R23, R24 \\
\hline
StableNormal & R02, R13 \\
\hline
SAM 3 & R01, R03, R04, R09, R11, R12, R15, R16, R21, R22, R23, R24 \\
\hline
SpatialTracker V2 & R05, R12, R15, R16, R17, R20, R21, R22, R23, R24 \\
\hline
Any6D & R10, R17 \\
\hline
FlowSeek & R19 \\
\toprule
\end{tabular}
\end{minipage}\hfill
\begin{minipage}[t]{0.47\textwidth}
\centering
\tiny
\setlength{\tabcolsep}{2pt}
\begin{tabular}{l >{\raggedright\arraybackslash}p{0.7\textwidth}}
\toprule
\multicolumn{1}{c}{\bf MODEL} & \multicolumn{1}{c}{\bf COVERED DIMENSIONS} \\
\hline
VFIMamba & R20 \\
\hline
ReasonX & R13, R18 \\
\hline
Qwen3.8-27B & R04, R06, R07, R10, R14, R25; IF-1 to IF-9, AC-1, CS-1 to CS-3 \\
\hline
DINOv3 & FQ-1, IF-10 \\
\hline
CLIP & TQ-4, IF-10 \\
\hline
FineVQ & FQ-2 \\
\hline
OpenCV/Farneback & TQ-1, TQ-2, TQ-3, TQ-5 \\
\toprule
\end{tabular}
\end{minipage}
\end{table}

\subsubsection{From Specialized Models to Physical Quantities}
\label{sec:method:quantities}

Where quantification is possible, each physical dimension is paired with a specialized vision foundation model that outputs numerical physical quantities rather than semantic judgments; this is the fundamental difference between PVBench and previous physical benchmarks. But not all physical quantities can be directly quantified, in which case MLLM-assisted judgment is introduced. By evaluation method, the physical evaluation of PVBench is thus divided into three paradigms: \textbf{\textit{i)}} \textbf{specialized-model quantitative evaluation}; \textbf{\textit{ii)}} \textbf{specialized-model measurement with MLLM judgment}; \textbf{\textit{iii)}} \textbf{pure MLLM semantic scoring}. Among the 25 physical dimensions, 17 are evaluated purely numerically, 4 rely on pure MLLM scoring, and 4 use hybrid measurement. 

\subsubsection{Score Mapping}
\label{sec:method:mapping}

Physical quantity values are mapped to a 1 to 5 score through thresholds, which are manually set according to the reasonable range of the physical quantity; their fit to human judgment is verified by the human validation in Section~\ref{sec:method:calibration}. Each dimension first takes the mean over its samples, then the domain and overall scores are obtained by equal-weight averaging across dimensions, avoiding dimensions with more samples from dominating the total. In the final report, scores are uniformly linearly mapped to a percentile scale.The specific physical quantities, formulas, specialized models used, and threshold-mapping details for each physical dimension are given in Appendix~\ref{sec:appendix:scoring}.

\subsection{Human Alignment For Benchmark Validation}
\label{sec:method:calibration}

To verify the consistency between evaluator scores and human judgments, we selected three T2V models and 5 evaluation prompts per physical dimension for manual annotation and scoring. For each evaluation dimension, we wrote detailed guidelines and examples, and annotators were trained first to fully understand the metric definition. Each sample was independently scored by 10 annotators and averaged. The results of this experiment are presented in Section~\ref{sec:exp:analysis}.

\subsection{Novelty: From Semantic Judgment to Physical-State Evaluation}
\label{sec:method:novelty}
PVBench differs from prior benchmarks primarily in the unit of evaluation: instead of asking whether a video looks physically plausible, it estimates observable physical states and evaluates their consistency with prompt-conditioned physical constraints and laws. This shift from holistic semantic judgment to computable physical-state evaluation is the core novelty of PVBench. In contrast to prior video benchmarks, which either emphasize semantic and perceptual quality or assess physical plausibility mainly through human or MLLM judgments, binary outcomes, proxy metrics, or camera-centric geometry, PVBench makes physical plausibility measurable at the level of object states and physical laws.

\section{Experiments}
\label{sec:experiments}

\subsection{Experimental Setup}
\label{sec:exp:setup}

\textbf{Evaluated models.} We evaluate 6 mainstream video generation models, covering open-source and closed-source systems: MiniMax H3~\citep{minimax2026h3}, SANA-Video 2.0~\citep{chen2026sanavideo2}, Wan2.1~\citep{wang2025wan} and Wan2.2~\citep{wang2025wan}, Seedance 2.5~\citep{seedance2026seedance2.5,seed2026seedance2}, and LingBot-Video~\citep{ma2026lingbot}, for a total of 6 T2V and 6 I2V configurations. The specific list is in Table~\ref{tab:leaderboard}.

\textbf{Generation protocol.} Each model generates all 600 prompt groups one by one; T2V and I2V variants use text and reference-image conditions respectively. All models use a unified output specification: 720p resolution, 10-second duration, and a fixed random seed. Full details of the generation parameters are provided in Appendix~\ref{sec:appendix:protocol}.

\textbf{Evaluation protocol.} The generated videos are uniformly evaluated according to the protocol in Section~\ref{sec:method:protocol}; all evaluators use fixed versions and weights to ensure reproducibility of the experiment. Final scores are uniformly reported on a percentile scale.

\subsection{Main Results and Cross-Layer Model Behavior}
\label{sec:exp:main}

\begin{figure*}[!t]
  \centering
  \includegraphics[width=\textwidth,trim={90bp 270bp 10bp 135bp},clip]{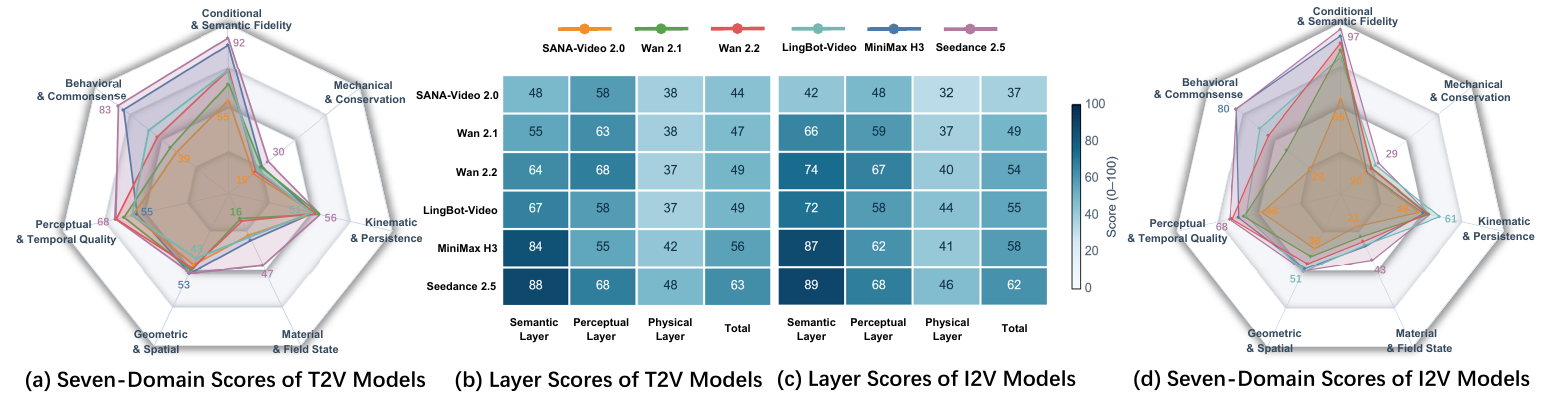}
  \caption{\textbf{PVBench results for six T2V and six I2V configurations.} The figure presents seven-domain radar plots for T2V and I2V together with heatmaps of semantic-, perceptual-, and physical-layer scores and the overall score, rounded to integers. Physical-layer scores are markedly lower than semantic- and perceptual-layer scores; Seedance 2.5 performs best overall, followed by MiniMax H3.}
  \label{fig:main_results}
\end{figure*}

Table~\ref{tab:leaderboard} summarizes the current percentile scores of the 6 T2V and 6 I2V models across all seven domains. Different models show clear capability differences across the seven metrics. Fig.~\ref{fig:main_results} further visualizes the score distribution of each model across the seven domains and the three-layer structure with radar and bar charts.

\begin{table}[!htbp]
\caption{\textbf{PVBench scores for the evaluated models.} Scores are reported on a percentile scale. Within each configuration, the highest score for each metric is boldfaced and the second-highest is marked with an asterisk. Open- and closed-source models are grouped separately; closed-source model rows are shaded light blue.}
\label{tab:leaderboard}
\vspace{-10pt}
\begin{center}
\scriptsize
\setlength{\tabcolsep}{2pt}
\renewcommand{\arraystretch}{1.05}
\begin{tabular}{@{}ccccccccc@{}}
\toprule
\multirow{2}{*}{\bf Config.} & \multirow{2}{*}{\bf Model} & \multicolumn{2}{c}{\bf Semantic Layer} & \multicolumn{1}{c}{\bf Perceptual Layer} & \multicolumn{4}{c}{\bf Physical Layer} \\
\cmidrule(lr){3-4}\cmidrule(lr){5-5}\cmidrule(lr){6-9}
 & & \multicolumn{1}{c}{\bf CSF} & \multicolumn{1}{c}{\bf BC} & \multicolumn{1}{c}{\bf PTQ} & \multicolumn{1}{c}{\bf GS} & \multicolumn{1}{c}{\bf MFS} & \multicolumn{1}{c}{\bf KP} & \multicolumn{1}{c}{\bf MC} \\
\hline
\multirow{6}{*}{T2V} & SANA-Video 2.0 & 55.1 & 39.0 & 58.2 & 47.0 & 27.4 & 52.2 & 18.7 \\
 & Wan2.1 & 64.8 & 43.7 & 63.4 & 50.6 & 16.4 & 55.3$^{*}$ & 24.8 \\
 & Wan2.2 & 73.1 & 53.6 & 68.1$^{*}$ & 49.2 & 18.1 & 52.9 & 20.5 \\
 & LingBot-Video & 72.8 & 59.8 & 58.0 & 42.8 & 28.9 & 51.1 & 22.5 \\
 & MiniMax H3 & 88.0$^{*}$ & 79.3$^{*}$ & 55.4 & \textbf{52.6} & 30.8$^{*}$ & 51.5 & 25.3$^{*}$ \\
\rowcolor{blue!8} & Seedance 2.5 & \textbf{92.2} & \textbf{83.3} & \textbf{68.4} & 52.2$^{*}$ & \textbf{47.2} & \textbf{55.6} & \textbf{30.0} \\
\hline
\multirow{6}{*}{I2V} & SANA-Video 2.0 & 56.0 & 22.9 & 48.4 & 35.8 & 21.3 & 48.0 & 19.5 \\
 & Wan2.1 & 84.6 & 41.1 & 59.2 & 40.9 & 28.0 & 53.9 & 24.0 \\
 & Wan2.2 & 88.6 & 55.1 & 66.6$^{*}$ & 45.8 & 31.1 & 52.6 & 23.6 \\
 & LingBot-Video & 80.2 & 61.8 & 58.1 & \textbf{50.8} & 33.4 & \textbf{60.7} & 26.8$^{*}$ \\
 & MiniMax H3 & 92.8$^{*}$ & \textbf{79.9} & 62.5 & 48.9 & 34.2$^{*}$ & 50.7 & 20.1 \\
\rowcolor{blue!8} & Seedance 2.5 & \textbf{96.8} & 79.7$^{*}$ & \textbf{67.7} & 50.3$^{*}$ & \textbf{43.4} & 54.4$^{*}$ & \textbf{29.0} \\
\toprule
\end{tabular}
\end{center}
\end{table}

\noindent\textbf{Semantic and Perceptual Layer Results.}
Overall, the semantic and perceptual layers can distinguish different models in condition following, behavior understanding, and visual quality, but the three do not show a completely consistent ranking. For example, SANA-Video 2.0 in the T2V configuration scores 58.2 in perception and temporal quality (PTQ), but only 39.0 in behavioral and causal commonsense (BC). This illustrates that a model can produce reasonably polished frames and temporal dynamics while remaining weak in action causality and commonsense reasoning; perceptual quality and behavioral plausibility are not equivalent. The semantic and perceptual layer results provide a reference for the physical-layer analysis.

\noindent\textbf{Physical Layer Results.}
Overall, the scores of geometric space and kinematic persistence are higher than those of material field states and mechanical conservation, indicating that generated videos maintain visible geometric structure and general motion trajectories relatively stably, while material field states and mechanical/conservation relations are more difficult. The low-score distribution of mechanical conservation especially shows that visual continuity or action completion alone is insufficient to guarantee that the result conforms to mechanical laws.

\insightbox{1}{Visual quality is not equivalent to physical plausibility}{Although mainstream models can score highly on semantic and perceptual metrics, their physical-layer scores are generally lower. Some models that lead in visual quality and semantic fidelity rank lower physically, showing that visual impression or semantic alignment alone does not establish physical plausibility.}

\insightbox{2}{Models remain weak at mechanics and conservation laws}{Unlike geometry and general motion, mechanical conservation jointly involves object trajectories, interactions, and energy or momentum relations, demanding stronger spatiotemporal modeling. Scores remain low across models, revealing deficiencies in collisions, forces, and energy changes.}

\subsection{Supplementary Analysis}
\label{sec:exp:analysis}

\noindent\textbf{Embodied Samples and High-Difficulty Supplemental Subset.}
We merge prompts tagged with embodied agents, occlusion, visual clutter, or camera motion into a high-difficulty supplemental subset, and compare it with the normal subset without such tags. Table~\ref{tab:confound} reports the physical-layer macro-average scores of the two subsets. Among T2V, 5 of 6 models drop on the supplemental subset, with an average relative drop of about 7.1\%; among I2V, 5 of 6 models drop, with an average relative drop of about 10.9\%.

\begin{figure*}[!htbp]
\centering
\begin{minipage}[t]{0.49\textwidth}
\captionsetup{width=\linewidth,justification=raggedright,singlelinecheck=false,skip=2pt}
\centering
\fontsize{5.5}{6.5}\selectfont
\setlength{\tabcolsep}{1.8pt}
\renewcommand{\arraystretch}{1.0}
\captionof{table}{\textbf{Physical-layer scores on normal and high-difficulty subsets.}}
\label{tab:confound}
\resizebox{\linewidth}{!}{%
\begin{tabular}{@{}ccccc@{}}
\toprule
\multirow{2}{*}{\bf Config.} & \multirow{2}{*}{\bf Model} & \multicolumn{1}{c}{\bf Normal} & \multicolumn{1}{c}{\bf High-Difficulty} & \multirow{2}{*}{\bf Difference} \\
 & & \multicolumn{1}{c}{\bf Score} & \multicolumn{1}{c}{\bf Score} & \\
\hline
\multirow{6}{*}{T2V} & MiniMax H3 & 42.6 & 40.2 & -2.4 \\
 & SANA-Video 2.0 & 38.6 & 37.1 & -1.5 \\
 & Wan2.1 & 38.6 & 35.4 & -3.3 \\
 & Wan2.2 & 37.1 & 35.8 & -1.3 \\
 & LingBot-Video & 39.1 & 33.6 & -5.5 \\
\rowcolor{blue!8} & Seedance 2.5 & 47.7 & 50.0 & +2.3 \\
\hline
\multirow{6}{*}{I2V} & MiniMax H3 & 40.4 & 41.3 & +0.9 \\
 & SANA-Video 2.0 & 33.3 & 26.6 & -6.6 \\
 & Wan2.1 & 37.9 & 34.3 & -3.6 \\
 & Wan2.2 & 40.9 & 35.1 & -5.7 \\
 & LingBot-Video & 44.0 & 43.4 & -0.6 \\
\rowcolor{blue!8} & Seedance 2.5 & 47.6 & 42.7 & -4.8 \\
\toprule
\end{tabular}
}%
\end{minipage}\hfill
\begin{minipage}[t]{0.49\textwidth}
\captionsetup{width=0.88\linewidth,justification=raggedright,singlelinecheck=false}
\centering
\tiny
\setlength{\tabcolsep}{2pt}
\captionof{table}{\textbf{Correlation between automatic scores and human ratings by domain.}}
\label{tab:human}
\vspace{-8pt}
\resizebox{0.82\linewidth}{!}{%
\begin{tabular}{c|ccccc}
\toprule
\multicolumn{1}{c}{\bf Metric} & \multicolumn{1}{c}{\bf GS} & \multicolumn{1}{c}{\bf MFS} & \multicolumn{1}{c}{\bf KP} & \multicolumn{1}{c}{\bf MC} & \multicolumn{1}{c}{\bf Total} \\
\hline
Spearman $\rho$ & 0.660 & 0.805 & 0.732 & 0.782 & 0.772 \\
Kendall $\tau$ & 0.528 & 0.692 & 0.619 & 0.651 & 0.643 \\
\toprule
\end{tabular}
}%
\par\vspace{0em}
\scriptsize
\setlength{\tabcolsep}{2pt}\renewcommand{\arraystretch}{1.0}
\captionsetup{width=0.82\linewidth,justification=raggedright,singlelinecheck=false}
\captionof{table}{\textbf{MiniMax H3: T2V vs. I2V.}}
\label{tab:t2vi2v}
\vspace*{-2pt}
\renewcommand{\arraystretch}{1.0}
\begin{tabular*}{0.82\linewidth}{@{\extracolsep{\fill}}crrc@{}}
\toprule
\multicolumn{1}{c}{\bf Domain} & \multicolumn{1}{c}{\bf T2V} & \multicolumn{1}{c}{\bf I2V} & \multicolumn{1}{c}{\bf Difference} \\
\hline
CS & 88.0 & 92.8 & +4.8 \\
BC & 79.3 & 79.9 & +0.6 \\
PTQ & 55.4 & 62.5 & +7.1 \\
GS & 52.6 & 48.9 & -3.7 \\
MFS & 30.8 & 34.2 & +3.4 \\
KP & 51.5 & 50.7 & -0.8 \\
MC & 25.3 & 20.1 & -5.2 \\
\toprule
\end{tabular*}\par\vspace{0.35em}
\end{minipage}
\end{figure*}

\insightbox{3}{The physical performance of most models drops in complex scenes}{Complex scene factors increase the burden of target retention, structure recovery, and dynamic reasoning, although models differ in their sensitivity to these factors.}

\noindent\textbf{Human Alignment Validation.}
We used 3 T2V models, 25 physical dimensions, and 5 prompts per dimension for human validation. Each sample was independently scored by 10 annotators and averaged; the inter-annotator agreement of the average scores from the 10 annotators is high (ICC(2,k) = 0.89)\citep{shrout1979intraclass}. Table~\ref{tab:human} reports the correlation between automatic evaluation scores and human scores by the four physical-layer domains (three models merged). The Spearman correlation coefficients\citep{spearman1904proof} of the four domains are between 0.660 and 0.805 (total 0.772), and the Kendall coefficients\citep{kendall1938new} are between 0.528 and 0.692 (total 0.643), indicating that the automatic evaluation results maintain generally high consistency with human judgments in the physical layer. Per-dimension human-alignment results are reported in Appendix~\ref{sec:appendix:perdim}.

\begin{figure}[!htbp]
  \centering
  \includegraphics[width=\linewidth,trim={16.6bp 129.2bp 13.0bp 88.9bp},clip]{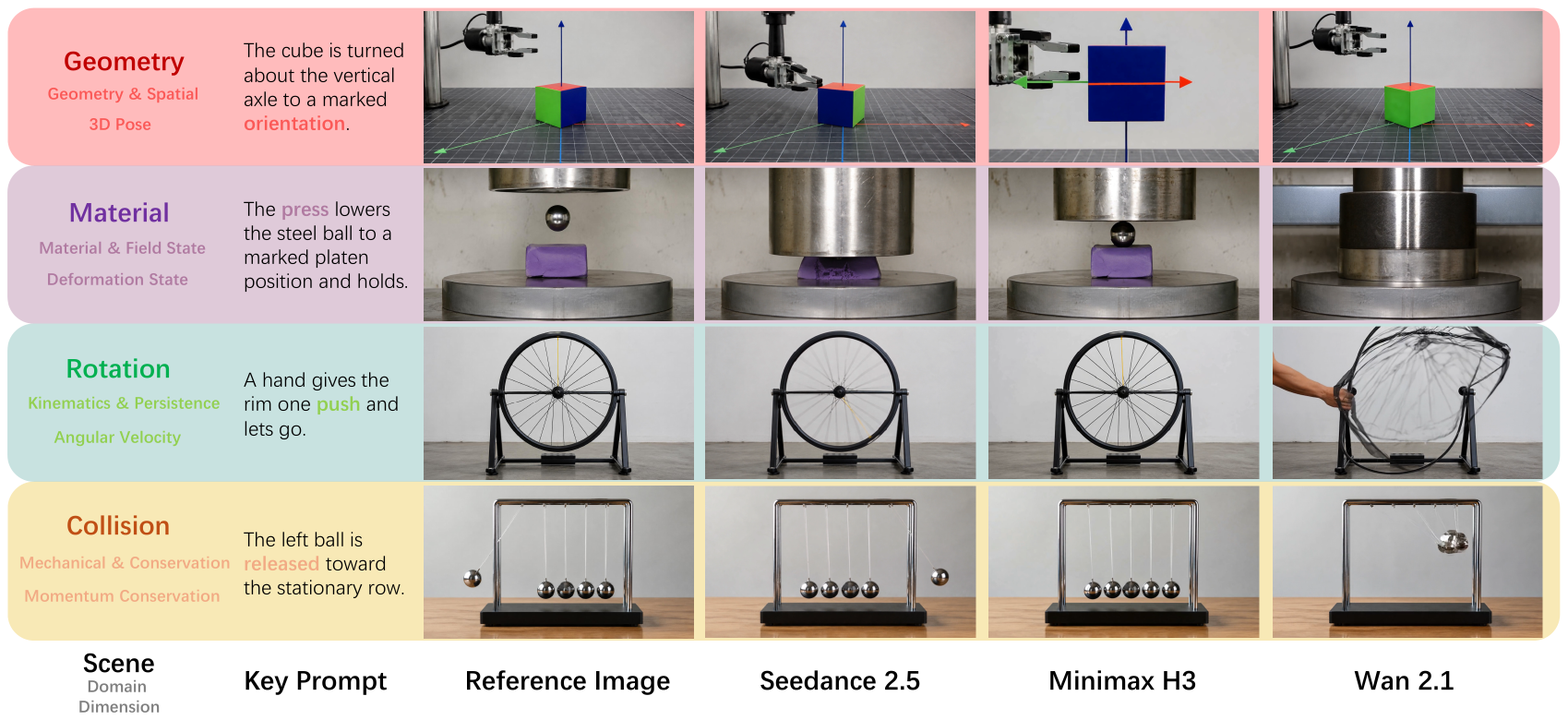}
  \caption{\textbf{Visual Comparison of Selected I2V Models.} Qualitative comparison of representative I2V generation results across selected models, highlighting differences in physical plausibility (e.g., geometry, material, motion, and conservation).}
  \label{fig:i2v_qualitative}
\end{figure}

\noindent\textbf{T2V \textit{vs.} I2V: the Role of First-Frame Prior.}
To analyze the role of the first-frame prior, we first compare the T2V and I2V configurations of MiniMax H3. Table~\ref{tab:t2vi2v} shows that I2V scores higher than T2V in both the semantic and perceptual layers, but the physical-layer results do not improve synchronously (Fig.~\ref{fig:i2v_qualitative}). I2V improves by 3.4 percentage points in material field states and is roughly flat in kinematic persistence, but drops by 3.7 and 5.2 percentage points in geometric space and mechanical conservation respectively. Per dimension, I2V improves more noticeably in light-shadow temporal consistency and magnetic-electric interaction, while depth, energy conservation, and object permanence scores drop. Thus, the first-frame prior does not affect different physical dimensions consistently. We report the paired differences for all six models in Appendix~\ref{sec:appendix:paired}.

\insightbox{4}{The first-frame prior mainly improves condition following and perceptual quality, not all physical quantities}{The first-frame prior can constrain the appearance of the initial scene and condition following, but cannot guarantee that subsequent geometry, motion, and mechanical relations remain physically reasonable.}

\subsection{Ablation Studies for {\bench}}
\label{sec:exp:ablation}

\begin{figure*}[!t]
\centering
\begin{minipage}[t]{0.43\textwidth}
\centering
\footnotesize
\setlength{\tabcolsep}{2pt}
\captionof{table}{\textbf{Pairwise consistency across VLM backends.}}
\label{tab:vlm}
\vspace*{-5pt}
\renewcommand{\arraystretch}{1.2}
\begin{tabular}{cccc}
\toprule
\multicolumn{1}{c}{\bf Backend} & \multicolumn{1}{c}{\bf Qwen3.5} & \multicolumn{1}{c}{\bf Qwen3.8} & \multicolumn{1}{c}{\bf InternVL3.5} \\
\hline
Qwen3.5 & 1.00 & 0.836 & 0.774 \\
Qwen3.8 & 0.836 & 1.00 & 0.705 \\
InternVL3.5 & 0.774 & 0.705 & 1.00 \\
\toprule
\end{tabular}\par\vspace{0.7em}
\end{minipage}\hfill
\begin{minipage}[t]{0.51\textwidth}
\centering
\footnotesize
\setlength{\tabcolsep}{2pt}
\captionof{table}{\textbf{Discriminability and human agreement of physical-quantity and MLLM scores.}}
\label{tab:numsem}
\vspace*{-5pt}
\resizebox{\linewidth}{!}{%
\begin{tabular}{>{\centering\arraybackslash}m{1.65in}cc}
\toprule
\multicolumn{1}{>{\centering\arraybackslash}m{1.65in}}{\bf Metric} & \multicolumn{1}{c}{\bf \shortstack{Physical-quantity\\score}} & \multicolumn{1}{c}{\bf \shortstack{MLLM\\score}} \\
\hline
Global entropy & 2.25 & 1.12 \\
Inter-dimension variance & 1.05 & 0.14 \\
Full-mark ratio & 15.5\% & 73.7\% \\
Human Spearman $\rho$ & 0.84 & 0.14 \\
\toprule
\end{tabular}
}%
\end{minipage}
\end{figure*}

\noindent\textbf{Different VLM Backends.}
We replace the VLM backend of the evaluator with Qwen3.5-9B~\citep{qwen2026qwen3.5}, Qwen3.8-27B~\citep{qwen2026qwen3.8}, or InternVL3.5-8B~\citep{wang2025internvl3.5}, and compare the consistency of their scores. The pairwise Spearman correlations range from 0.71 to 0.84, indicating high agreement. This suggests that the evaluation is robust to the choice of VLM backend. We use Qwen3.8-27B by default.

\noindent\textbf{Discriminability: Numerical Physical Quantities vs.\ MLLM Semantic Scores.}
For the same physical dimensions, we compare scores from specialized numerical models with direct MLLM scores. Numerical scores have higher global entropy (2.25 vs. 1.12) and inter-dimension variance (1.05 vs. 0.14). MLLM scores are often saturated, with 73.7\% at the maximum. Against human ratings, numerical scores reach a Spearman correlation of 0.84, compared with 0.14 for MLLM scores. These results show that physical-quantity scoring is both more discriminative and more aligned with human judgments (Table~\ref{tab:numsem}).

\insightbox{5}{Physical-quantity computation provides a more reliable evaluation signal than semantic scoring}{Unlike MLLM semantic scores, which tend to saturate and diverge from human judgment, numerical physical quantities offer a more discriminative and human-aligned basis for assessing physical plausibility. This suggests that future benchmarks should prioritize computable physical measurements over language-based semantic judgments.}

\begin{wraptable}{r}{0.48\textwidth}
\vspace{-5pt}
\centering
\captionsetup{width=\linewidth,justification=raggedright,singlelinecheck=false}
\caption{\textbf{Evaluator threshold calibration.} Default thresholds are boldfaced.}
\label{tab:calib}
\vspace{-10pt}
\scriptsize
\setlength{\tabcolsep}{1.5pt}
\renewcommand{\arraystretch}{1.15}
\begin{tabular}{cccc}
\toprule
\makecell{\bf Evaluator\\\bf hyperparameter} & \multicolumn{1}{c}{\bf Threshold} & \multicolumn{1}{c}{\bf Spearman $\rho$} & \multicolumn{1}{c}{\bf Coverage} \\
\hline
\multirow{3}{*}{\makecell{SAM3\\confidence}} & 0.05 & 0.490 & 84.3\% \\
 & \textbf{0.10} & \textbf{0.601} & \textbf{84.3\%} \\
 & 0.50 & 0.132 & 79.8\% \\
\cline{2-4}
\multirow{3}{*}{\makecell{Trajectory\\object area}} & 0.05\% & 0.733 & 99.2\% \\
 & \textbf{0.10\%} & \textbf{0.858} & \textbf{99.2\%} \\
 & 0.50\% & 0.666 & 99.2\% \\
\bottomrule
\end{tabular}
\vspace{-10pt}
\end{wraptable}

\noindent\textbf{Threshold Calibration.}
We calibrate two key evaluator thresholds: SAM3~\citep{metasam2025sam3} detection confidence and trajectory object area. As shown in Table~\ref{tab:calib}, the PVBench thresholds of 0.10 and 0.1\% perform best in both coverage and Spearman correlation with human scores, whereas 0.50 and 0.50\% perform worst on both measures.

\noindent\textbf{Physical-Layer Domain Ablation.}
To examine how each physical-layer domain contributes to the overall physical score, we perform a leave-one-out ablation on the four domains (GS, MFS, KP, MC) using the results in Table~\ref{tab:leaderboard}. For each model, we recompute the physical-layer average after removing one domain. Averaged over the six T2V models, the full physical-layer score is 38.5. Removing MC raises the average to 43.4, and removing MFS raises it to 41.9, whereas removing GS or KP decreases it to 35.0 and 33.6, respectively. This shows that MC and MFS are the main factors pulling down the overall physical score, while GS and KP contribute relatively higher scores. The consistently low MC scores further indicate that current models are generally weak in mechanics and conservation laws. These results support reporting the four physical domains separately, as they reveal distinct weaknesses across different aspects of physical plausibility.

\FloatBarrier

\section{Conclusion and Future Work}
\label{sec:conclusion}

Video generation looks increasingly realistic, yet its physical plausibility remains poorly measured. Existing evaluations rely on semantic binary judgments or camera-centric geometry, and cannot reduce physical plausibility to computable, testable quantities. We propose PVBench, a physical-quantity-driven benchmark that extracts numerical measurements with specialized vision models, checks them against physical laws, uses same-scene T2V/I2V paired prompts, and resorts to MLLM scoring only for hard-to-quantify dimensions. Our key finding is that numerical physical-quantity scores are more discriminative than MLLM semantic scores and highly consistent with human judgments, enabling more reliable and quantifiable evaluation. We hope PVBench becomes a quantifiable yardstick for physically plausible video generation.

\noindent\textbf{Limitations and Future Work.} The evaluation depends on specialized vision models and skips only a small number of samples for unobservable cases; material and phase-change dimensions still use MLLM scoring. We only ablate the VLM backend, and a full ablation of the specialized evaluators is left for future work. Moreover, the current evaluation largely relies on checking consistency with physical laws and, except for a few dimensions, lacks ground-truth references. More robust tracking and finer physical quantities will improve coverage. In future work, we plan to build ground-truth references for conditional image-to-video (CI2V) models using physics engines and other simulation tools to enable more direct evaluation. We will also expand the benchmark as compute and open-source models grow.

\clearpage
\section*{AI Use Statement}
\label{sec:aiuse}
We used generative AI tools to polish the writing and to assist in constructing the prompt set. All AI-assisted content was reviewed and verified by the authors. We take full responsibility for the final content, claims, and artifacts in this work.

\section*{Ethics Statement}
\label{sec:ethics}
We have read and will follow the ICLR Code of Ethics. The human study was limited to validating the automatic evaluators. Annotators were trained with detailed guidelines and examples, and each sample was independently scored by 10 annotators and averaged. The annotation followed the protocol described in this paper.

\section*{Reproducibility Statement}
\label{sec:repro}
We describe the benchmark, prompts, evaluators, scoring rules, and experimental protocols in the main paper and appendix. During evaluation, we fix model versions, weights, and random seeds. We will release the code, prompt set, and calibration protocol to support reproduction and future comparisons. The experiments cover 12 model configurations and 600 prompt groups, about 7,200 generated videos, evaluated on a single H20 GPU.

\bibliography{pvbench}
\appendix
\setcounter{table}{0}
\setcounter{figure}{0}
\renewcommand{\thetable}{A\arabic{table}}
\renewcommand{\thefigure}{A\arabic{figure}}
\renewcommand{\theHtable}{A\arabic{table}}
\renewcommand{\theHfigure}{A\arabic{figure}}
\clearpage
\section*{Appendix}

\subsection*{Appendix Overview}
The supplementary material provides detailed benchmark specifications, scoring procedures, and results to support reproducibility and facilitate future comparisons:
\begin{itemize}
  \item Appendix~\ref{sec:appendix:dimcat} describes the 46 evaluation dimensions and their organization, and compares the physical-dimension coverage of PVBench with prior benchmarks.
  \item Appendix~\ref{sec:appendix:perdim} reports per-dimension human-alignment results for the physical dimensions.
  \item Appendix~\ref{sec:appendix:alldim} provides the per-dimension automatic evaluation scores for all 12 model configurations.
  \item Appendix~\ref{sec:appendix:scoring} details the scoring principles, including the specialized models and evaluation paradigms, physical quantities and formulas, threshold mappings, and semantic/perceptual scoring criteria.
  \item Appendix~\ref{sec:appendix:promptstats} gives additional statistics for the 600 prompt groups.
  \item Appendix~\ref{sec:appendix:protocol} specifies the video-generation protocol and model configurations.
  \item Appendix~\ref{sec:appendix:repro} reports evaluation costs and reproducibility details.
\end{itemize}

\section{Evaluation Dimensions: Description and Master Table}
\label{sec:appendix:dimcat}

\subsection{Dimension Description}
\label{sec:appendix:dimcat:desc}
IF-10 (reference condition following) applies only to I2V conditioned on a reference image; T2V does not evaluate this dimension. Therefore T2V evaluates 45 dimensions in total, and I2V evaluates all 46 dimensions. Aggregation is computed by equal weight over the dimensions actually evaluated in each modality.

\subsection{Master Table of Evaluation Dimensions}
\label{sec:appendix:catalog}

Table~\ref{tab:dimcat} gives the ID, name, domain (D1 to D7), and assessment layer (L1 semantic layer / L2 perceptual layer / L3 physical layer) of all 46 PVBench evaluation dimensions. Per-dimension human-alignment results for the physical layer are reported in Appendix~\ref{sec:appendix:perdim}, Table~\ref{tab:perdim}. Per-dimension automatic scores for all 46 dimensions across the 12 model configurations are reported in Appendix~\ref{sec:appendix:alldim}, Tables~\ref{tab:alldim_d1} through \ref{tab:alldim_d7}. The physical-dimension coverage comparison with prior benchmarks is in this section, Table~\ref{tab:coverage}.

\begin{table}[!htbp]
\caption{Master table of all 46 PVBench evaluation dimensions.}
\label{tab:dimcat}
\begin{center}
\small
\begin{tabular}{lccc}
\toprule
\multicolumn{1}{c}{\bf DIM} & \multicolumn{1}{c}{\bf NAME} & \multicolumn{1}{c}{\bf DOMAIN} & \multicolumn{1}{c}{\bf LAYER} \\
\hline
IF-1 & Subject adherence & D1 & L1 \\
IF-2 & Attribute adherence & D1 & L1 \\
IF-4 & Spatial relation adherence & D1 & L1 \\
IF-5 & Style adherence & D1 & L1 \\
IF-7 & Negation adherence & D1 & L1 \\
IF-8 & Scene matching & D1 & L1 \\
IF-9 & Camera motion adherence & D1 & L1 \\
IF-10 & Reference condition fidelity & D1 & L1 \\
IF-3 & Action adherence & D2 & L1 \\
IF-6 & Temporal adherence & D2 & L1 \\
AC-1 & Action causality & D2 & L1 \\
CS-1 & Physical commonsense & D2 & L1 \\
CS-2 & Biological commonsense & D2 & L1 \\
CS-3 & Social commonsense & D2 & L1 \\
FQ-1 & Subject consistency & D3 & L2 \\
FQ-2 & Aesthetic quality & D3 & L2 \\
TQ-1 & Motion smoothness & D3 & L2 \\
TQ-2 & Temporal flickering & D3 & L2 \\
TQ-3 & Dynamic degree & D3 & L2 \\
TQ-4 & Background consistency & D3 & L2 \\
TQ-5 & Imaging quality & D3 & L2 \\
R01 & Depth & D4 & L3 \\
R02 & Surface normal & D4 & L3 \\
R03 & Shape & D4 & L3 \\
R04 & Absolute scale & D4 & L3 \\
R08 & Multi-view geometric consistency & D4 & L3 \\
R09 & Relative scale & D4 & L3 \\
R10 & 3D pose & D4 & L3 \\
R11 & Spatial relation & D4 & L3 \\
R19 & Cross-frame geometric consistency & D4 & L3 \\
R05 & Current deformation state & D5 & L3 \\
R06 & Material properties & D5 & L3 \\
R07 & Thermal radiation & D5 & L3 \\
R13 & Light and shadow spatial consistency & D5 & L3 \\
R14 & Magnetic and electric interaction & D5 & L3 \\
R18 & Light and shadow temporal consistency & D5 & L3 \\
R25 & Phase transition & D5 & L3 \\
R15 & Displacement & D6 & L3 \\
R16 & Velocity \& acceleration & D6 & L3 \\
R17 & Angular velocity & D6 & L3 \\
R20 & Motion continuity & D6 & L3 \\
R21 & Object permanence & D6 & L3 \\
R12 & Stability & D7 & L3 \\
R22 & Dynamic effects of forces & D7 & L3 \\
R23 & Momentum conservation & D7 & L3 \\
R24 & Energy conservation & D7 & L3 \\
\toprule

\end{tabular}
\end{center}
\end{table}

\subsection{Physical Dimension Coverage vs.\ Prior Work}
\label{sec:appendix:coverage}

Table~\ref{tab:coverage} shows the per-dimension coverage comparison of PVBench with existing physical-plausibility benchmarks on the 25 physical dimensions. Existing benchmarks cover at most a small subset, and mainly through semantic judgment rather than physical-quantity computation; PVBench provides numerical evaluation on all three classes of physical dimensions.

\begin{table}[!htbp]
\caption{Per-dimension coverage of the 25 physical dimensions across physical-plausibility benchmarks. PVBench covers all three classes of physical dimensions; existing benchmarks cover at most a small subset, and predominantly via semantic judgments rather than numerical physical quantities. Column abbreviations: VP-2 = VideoPhy-2, PGB = PhyGenBench, P-IQ = Physics-IQ, PWB = PhyWorldBench, WS = WorldScore, WMB = WorldModelBench.}
\label{tab:coverage}
\begin{center}
\small
\resizebox{\linewidth}{!}{%
\begin{tabular}{l|ccccccc}
\toprule
\multicolumn{1}{c}{\bf DIMENSION} & \multicolumn{1}{c}{\bf OURS} & \multicolumn{1}{c}{\bf VP-2} & \multicolumn{1}{c}{\bf PGB} & \multicolumn{1}{c}{\bf P-IQ} & \multicolumn{1}{c}{\bf PWB} & \multicolumn{1}{c}{\bf WS} & \multicolumn{1}{c}{\bf WMB} \\
\hline \\
\multicolumn{8}{l}{\emph{Class I: Object-intrinsic physical properties}} \\
R01 Depth ordering & \cmark & \xmark & \xmark & \xmark & \xmark & \xmark & \xmark \\
R02 Surface normals & \cmark & \xmark & \xmark & \xmark & \xmark & \xmark & \xmark \\
R03 Shape & \cmark & \xmark & \xmark & \xmark & \xmark & \xmark & \xmark \\
R04 Absolute scale & \cmark & \xmark & \xmark & \xmark & \xmark & \xmark & \xmark \\
R05 Deformation state & \cmark & \cmark & \xmark & \cmark & \cmark & \xmark & \cmark \\
R06 Material properties & \cmark & \xmark & \cmark & \cmark & \cmark & \xmark & \cmark \\
R07 Thermal radiation & \cmark & \xmark & \xmark & \xmark & \xmark & \xmark & \xmark \\
\multicolumn{8}{l}{\emph{Class II: Object and space relations}} \\
R08 Multi-view geometric consistency & \cmark & \xmark & \xmark & \xmark & \xmark & \xmark & \xmark \\
R09 Relative scale & \cmark & \xmark & \xmark & \xmark & \cmark & \xmark & \xmark \\
R10 3D pose & \cmark & \xmark & \xmark & \xmark & \xmark & \xmark & \xmark \\
R11 Spatial relations & \cmark & \xmark & \xmark & \xmark & \xmark & \xmark & \cmark \\
R12 Stability & \cmark & \xmark & \xmark & \xmark & \cmark & \xmark & \xmark \\
R13 Light and shadow spatial consistency & \cmark & \cmark & \cmark & \cmark & \cmark & \xmark & \xmark \\
R14 Magnetic and electric interaction & \cmark & \xmark & \xmark & \cmark & \cmark & \xmark & \xmark \\
\multicolumn{8}{l}{\emph{Class III: Object dynamics across space and time}} \\
R15 Displacement & \cmark & \xmark & \xmark & \xmark & \xmark & \xmark & \xmark \\
R16 Velocity \& acceleration & \cmark & \xmark & \xmark & \xmark & \xmark & \xmark & \xmark \\
R17 Angular velocity & \cmark & \xmark & \xmark & \xmark & \xmark & \xmark & \xmark \\
R18 Light and shadow temporal consistency & \cmark & \xmark & \xmark & \xmark & \cmark & \xmark & \xmark \\
R19 Cross-frame geometric consistency & \cmark & \xmark & \xmark & \xmark & \xmark & \xmark & \xmark \\
R20 Motion continuity & \cmark & \xmark & \xmark & \xmark & \xmark & \cmark & \xmark \\
R21 Object permanence & \cmark & \xmark & \xmark & \xmark & \xmark & \xmark & \xmark \\
R22 Dynamic effects of forces & \cmark & \cmark & \cmark & \cmark & \cmark & \xmark & \cmark \\
R23 Momentum conservation & \cmark & \cmark & \xmark & \xmark & \xmark & \xmark & \xmark \\
R24 Energy conservation & \cmark & \xmark & \xmark & \xmark & \cmark & \xmark & \xmark \\
R25 Phase transition & \cmark & \xmark & \cmark & \cmark & \cmark & \xmark & \xmark \\
\toprule
\end{tabular}%
}
\end{center}
\end{table}

Table~\ref{tab:compare} summarizes the scope and evaluation approaches of representative physical-plausibility benchmarks. Coverage counts for prior benchmarks are derived from the per-dimension comparison in Table~\ref{tab:coverage}.

\begin{table}[!htbp]
\caption{PVBench compared with existing physical-plausibility benchmarks. Counts indicate the number of covered dimensions among the 25 physical dimensions in Table~\ref{tab:coverage}, not the total number of evaluation dimensions of each benchmark.}
\label{tab:compare}
\begin{center}
\footnotesize
\setlength{\tabcolsep}{2pt}
\begin{tabular}{>{\centering\arraybackslash}m{0.95in} >{\centering\arraybackslash}m{0.35in} >{\centering\arraybackslash}m{1.0in} >{\centering\arraybackslash}m{1.2in} >{\centering\arraybackslash}m{0.8in}}
\toprule
\multicolumn{1}{c}{\bf Benchmark} & \multicolumn{1}{c}{\bf Physical dims.} & \multicolumn{1}{c}{\bf Judge type} & \multicolumn{1}{c}{\bf Numerical evaluation} & \multicolumn{1}{c}{\bf Setting} \\
\hline
VideoPhy-2 & 4 & Human + automated & No & T2V \\
PhyGenBench & 4 & VLM & No & T2V \\
Physics-IQ & 6 & Motion-mask metrics & Proxy metrics & Real-video continuation \\
PhyWorldBench & 10 & Human + MLLM & No & T2V \\
WorldModel\\Bench & 4 & Human + VLM & No & T2V \\
WorldScore & 1 & Geometric SLAM & Partial (camera-centric) & T2V/I2V \\
\hline
PVBench (ours) & 25 & Specialized models + MLLM + physical laws & Yes & T2V/I2V \\
\toprule
\end{tabular}
\end{center}
\end{table}

\section{Per-Dimension Human Alignment Results}
\label{sec:appendix:perdim}

Table~\ref{tab:perdim} reports the correlation with human scores for each of the 25 physical dimensions (3 T2V models and 15 samples per dimension). The overall Spearman correlation is $\rho=0.772$ and Kendall correlation is $\tau=0.643$, consistent with the main-text results. Some dimensions that depend on detection, tracking, or long-term consistency have lower correlations, mainly because of skipped or unobservable samples.

\begin{table}[!htbp]
\caption{Per-dimension correlation between automatic evaluation scores and human scores (3 T2V models merged, 15 paired samples per dimension; the auto-score mean is the 3-model merged mean on the 1 to 5 scale).}
\label{tab:perdim}
\begin{center}
\small
\resizebox{\linewidth}{!}{%
\begin{tabular}{l|cccccc}
\toprule
\multicolumn{1}{c}{\bf DIM} & \multicolumn{1}{c}{\bf DIMENSION} & \multicolumn{1}{c}{\bf n} & \multicolumn{1}{c}{\bf AUTO MEAN} & \multicolumn{1}{c}{\bf Spearman $\rho$} & \multicolumn{1}{c}{\bf Kendall $\tau$} \\
\hline
R01 & Depth & 15 & 3.67 & 0.721 & 0.613 \\
R02 & Surface normals & 15 & 2.87 & 0.706 & 0.610 \\
R03 & Shape & 15 & 4.07 & 0.540 & 0.457 \\
R04 & Absolute scale & 15 & 1.93 & 0.638 & 0.532 \\
R05 & Deformation state & 15 & 2.53 & 0.622 & 0.563 \\
R06 & Material properties & 15 & 3.13 & 0.867 & 0.734 \\
R07 & Thermal radiation & 15 & 2.27 & 0.839 & 0.729 \\
R08 & Multi-view geometric consistency & 15 & 2.87 & 0.814 & 0.700 \\
R09 & Relative scale & 15 & 2.67 & 0.401 & 0.320 \\
R10 & 3D pose & 15 & 3.73 & 0.760 & 0.671 \\
R11 & Spatial relations & 15 & 1.93 & 0.693 & 0.619 \\
R12 & Stability & 15 & 1.73 & 0.777 & 0.679 \\
R13 & Light and shadow spatial consistency & 15 & 1.73 & 0.269 & 0.246 \\
R14 & Magnetic and electric interaction & 15 & 2.07 & 0.778 & 0.681 \\
R15 & Displacement & 15 & 4.47 & 0.765 & 0.676 \\
R16 & Velocity \& acceleration & 15 & 3.07 & 0.624 & 0.558 \\
R17 & Angular velocity & 15 & 1.27 & 0.440 & 0.388 \\
R18 & Light and shadow temporal consistency & 15 & 2.67 & 0.803 & 0.702 \\
R19 & Cross-frame geometric consistency & 15 & 3.40 & 0.687 & 0.599 \\
R20 & Motion continuity & 15 & 3.80 & 0.240 & 0.214 \\
R21 & Object permanence & 15 & 3.60 & 0.125 & 0.108 \\
R22 & Dynamic effects of forces & 15 & 2.67 & 0.591 & 0.507 \\
R23 & Momentum conservation & 15 & 2.13 & 0.640 & 0.567 \\
R24 & Energy conservation & 15 & 1.73 & 0.825 & 0.731 \\
R25 & Phase transition & 15 & 1.93 & 0.784 & 0.700 \\
\hline
Total & 25 phys.\ dims & 375 & N/A & 0.772 & 0.643 \\
\toprule
\end{tabular}%
}
\end{center}
\end{table}

\section{Per-Dimension Automatic Score Tables}
\label{sec:appendix:alldim}

Table~\ref{tab:skip} first reports the skipped dimension instances for each configuration. Tables~\ref{tab:alldim_d1} through \ref{tab:alldim_d7} then give the automatic evaluation mean score (1 to 5) of all 12 model configurations on all 46 PVBench dimensions, by the seven domains D1 to D7, obtained by averaging per dimension from the global report (semantic/perceptual layers, D1 to D3) and the unified physical report (physical layer, D4 to D7) in the result directory of each configuration; samples with status \emph{skipped} or \emph{error} (e.g., IF-10 not applicable to T2V reference images) are excluded from the mean and recorded as not applicable. Physical-layer dimensions are generally lower than semantic/perceptual layers, consistent with the insight in main text Section~\ref{sec:exp:main}.

\begin{table}[!htbp]
\caption{Skipped dimension instances for each model configuration. ``Total Dimension Evaluations'' denotes the sum of evaluation instances over all evaluated dimensions for each configuration. Normally this is 951. For Seedance 2.5 I2V, 69 dimension-level evaluation instances are missing because the corresponding videos failed to generate due to face-related content moderation, yielding 882 = 951 - 69. ``Skipped'' only counts instances skipped by the evaluator and does not include generation failures.}
\label{tab:skip}
\begin{center}
\scriptsize
\setlength{\tabcolsep}{2pt}
\begin{tabular}{llrrr}
\toprule
\bf Config. & \bf Model & \bf Skipped & \bf Total Instances & \bf Ratio \\
\hline
I2V & LingBot-Video & 46 & 951 & 4.84\% \\
T2V & LingBot-Video & 26 & 951 & 2.73\% \\
I2V & MiniMax H3 & 10 & 951 & 1.05\% \\
T2V & MiniMax H3 & 16 & 951 & 1.68\% \\
I2V & SANA-Video 2.0 & 24 & 951 & 2.52\% \\
T2V & SANA-Video 2.0 & 19 & 951 & 2.00\% \\
I2V & Seedance 2.5 & 15 & 882 & 1.70\% \\
T2V & Seedance 2.5 & 18 & 951 & 1.89\% \\
I2V & Wan2.1 & 15 & 951 & 1.58\% \\
T2V & Wan2.1 & 21 & 951 & 2.21\% \\
I2V & Wan2.2 & 16 & 951 & 1.68\% \\
T2V & Wan2.2 & 17 & 951 & 1.79\% \\
\toprule
\end{tabular}
\end{center}
\end{table}

\begin{table}[!htbp]
\caption{Automatic evaluation mean score (1 to 5) of all 12 configurations on D1 (instruction following and semantic alignment, 8 dims).}
\label{tab:alldim_d1}
\centering
\scriptsize
\begin{tabular}{l|cccccccc}
\toprule
\multicolumn{1}{c|}{\bf Config} & IF-1 & IF-2 & IF-4 & IF-5 & IF-7 & IF-8 & IF-9 & IF-10 \\
\hline
MiniMax H3 T2V & 4.89 & 4.80 & 4.81 & 4.65 & 3.29 & 5.00 & 4.19 &  \\
SANA-Video 2.0 T2V & 2.74 & 3.25 & 2.76 & 3.91 & 1.71 & 5.00 & 3.05 &  \\
Wan2.1-T2V-14B T2V & 3.63 & 3.75 & 4.05 & 4.09 & 2.38 & 4.95 & 2.29 &  \\
Wan2.2-T2V-A14B T2V & 3.89 & 4.80 & 4.43 & 3.91 & 2.33 & 5.00 & 3.10 &  \\
Seedance 2.5 T2V & 4.42 & 4.75 & 4.71 & 4.83 & 4.86 & 5.00 & 4.24 &  \\
LingBot-Video T2V & 4.00 & 3.30 & 3.76 & 3.22 & 4.86 & 5.00 & 3.24 &  \\
MiniMax H3 I2V & 4.68 & 4.85 & 4.76 & 5.00 & 5.00 & 5.00 & 4.19 & 4.20 \\
SANA-Video 2.0 I2V & 2.53 & 3.05 & 2.00 & 4.30 & 2.86 & 4.86 & 2.86 & 3.48 \\
Wan2.1-I2V-14B-720P I2V & 4.74 & 4.85 & 3.90 & 5.00 & 4.81 & 5.00 & 2.19 & 4.58 \\
Wan2.2-I2V-A14B I2V & 4.79 & 4.95 & 4.48 & 5.00 & 4.81 & 5.00 & 2.71 & 4.60 \\
Seedance 2.5 I2V & 4.95 & 5.00 & 4.80 & 5.00 & 5.00 & 5.00 & 4.56 & 4.67 \\
LingBot-Video I2V & 4.21 & 3.67 & 4.70 & 3.71 & 4.83 & 4.52 & 5.00 & 3.00 \\
\toprule
\end{tabular}%
\end{table}

\begin{table}[!htbp]
\caption{Automatic evaluation mean score (1 to 5) of all 12 configurations on D2 (action and commonsense reasoning, 6 dims).}
\label{tab:alldim_d2}
\centering
\scriptsize

\begin{tabular}{lcccccc}
\toprule
\multicolumn{1}{c|}{\bf Config} & IF-3 & IF-6 & AC-1 & CS-1 & CS-2 & CS-3 \\
\hline
MiniMax H3 T2V & 3.65 & 4.29 & 3.18 & 4.85 & 4.78 & 4.29 \\
SANA-Video 2.0 T2V & 1.60 & 1.76 & 1.94 & 3.70 & 3.26 & 3.10 \\
Wan2.1-T2V-14B T2V & 1.75 & 1.67 & 1.88 & 3.85 & 3.91 & 3.43 \\
Wan2.2-T2V-A14B T2V & 2.15 & 2.33 & 2.24 & 4.05 & 4.39 & 3.71 \\
Seedance 2.5 T2V & 4.00 & 4.29 & 3.18 & 4.95 & 4.87 & 4.71 \\
LingBot-Video T2V & 3.35 & 2.19 & 2.47 & 3.80 & 4.35 & 4.19 \\
MiniMax H3 I2V & 3.95 & 4.57 & 3.29 & 4.60 & 4.91 & 3.86 \\
SANA-Video 2.0 I2V & 1.45 & 1.86 & 1.47 & 2.80 & 1.87 & 2.05 \\
Wan2.1-I2V-14B-720P I2V & 1.70 & 2.10 & 2.18 & 3.75 & 3.39 & 2.76 \\
Wan2.2-I2V-A14B I2V & 2.25 & 2.48 & 2.12 & 4.35 & 4.61 & 3.43 \\
Seedance 2.5 I2V & 4.25 & 4.10 & 3.00 & 4.78 & 4.50 & 4.50 \\
LingBot-Video I2V & 3.45 & 3.38 & 2.71 & 4.05 & 3.87 & 3.38 \\
\toprule
\end{tabular}%
\end{table}

\begin{table}[!htbp]
\caption{Automatic evaluation mean score (1 to 5) of all 12 configurations on D3 (perception and quality, 7 dims).}
\label{tab:alldim_d3}
\centering
\scriptsize
\begin{tabular}{l|ccccccc}
\toprule
\multicolumn{1}{c|}{\bf Config} & FQ-1 & FQ-2 & TQ-1 & TQ-2 & TQ-3 & TQ-4 & TQ-5 \\
\hline
MiniMax H3 T2V & 3.76 & 3.36 & 1.18 & 3.74 & 4.33 & 3.00 & 3.13 \\
SANA-Video 2.0 T2V & 3.05 & 5.00 & 1.05 & 3.00 & 4.19 & 2.42 & 4.61 \\
Wan2.1-T2V-14B T2V & 2.81 & 3.86 & 2.64 & 4.70 & 4.29 & 3.38 & 3.09 \\
Wan2.2-T2V-A14B T2V & 3.81 & 4.45 & 2.27 & 4.43 & 4.33 & 3.38 & 3.39 \\
Seedance 2.5 T2V & 3.67 & 4.27 & 2.23 & 4.74 & 4.57 & 3.79 & 2.87 \\
LingBot-Video T2V & 2.19 & 3.00 & 2.41 & 4.57 & 4.67 & 3.50 & 2.91 \\
MiniMax H3 I2V & 3.00 & 3.91 & 2.05 & 4.09 & 4.52 & 3.75 & 3.17 \\
SANA-Video 2.0 I2V & 1.76 & 4.73 & 1.05 & 2.83 & 4.00 & 1.71 & 4.48 \\
Wan2.1-I2V-14B-720P I2V & 3.05 & 3.91 & 1.95 & 4.09 & 4.29 & 3.00 & 3.30 \\
Wan2.2-I2V-A14B I2V & 3.57 & 4.50 & 1.82 & 4.22 & 4.57 & 3.58 & 3.39 \\
Seedance 2.5 I2V & 3.80 & 4.05 & 2.15 & 4.75 & 4.31 & 3.86 & 3.04 \\
LingBot-Video I2V & 2.86 & 3.59 & 1.59 & 4.48 & 4.62 & 3.12 & 3.00 \\
\toprule
\end{tabular}%
\end{table}

\begin{table}[!htbp]
\caption{Automatic evaluation mean score (1 to 5) of all 12 configurations on D4 (geometry and space, 9 dims).}
\label{tab:alldim_d4}
\centering
\scriptsize
\begin{tabular}{lccccccccc}
\toprule
\multicolumn{1}{c|}{\bf Config} & R01 & R02 & R03 & R04 & R08 & R09 & R10 & R11 & R19 \\
\hline
MiniMax H3 T2V & 4.43 & 2.24 & 4.36 & 2.09 & 3.39 & 2.00 & 4.35 & 1.39 & 3.70 \\
SANA-Video 2.0 T2V & 3.59 & 2.38 & 3.18 & 1.38 & 3.65 & 3.27 & 3.40 & 1.33 & 3.74 \\
Wan2.1-T2V-14B T2V & 3.40 & 3.52 & 4.00 & 2.05 & 2.00 & 4.33 & 2.95 & 1.79 & 3.17 \\
Wan2.2-T2V-A14B T2V & 3.48 & 2.81 & 3.45 & 2.30 & 2.57 & 3.42 & 3.65 & 1.39 & 3.65 \\
Seedance 2.5 T2V & 3.67 & 2.76 & 4.36 & 2.41 & 3.26 & 2.12 & 4.25 & 1.67 & 3.30 \\
LingBot-Video T2V & 4.50 & 2.57 & 3.09 & 1.58 & 2.96 & 1.30 & 3.95 & 1.28 & 3.17 \\
MiniMax H3 I2V & 3.55 & 1.86 & 4.45 & 2.33 & 3.04 & 2.50 & 4.05 & 1.33 & 3.48 \\
SANA-Video 2.0 I2V & 3.00 & 2.33 & 2.09 & 2.14 & 3.17 & 2.40 & 2.10 & 1.22 & 3.43 \\
Wan2.1-I2V-14B-720P I2V & 3.48 & 1.95 & 3.27 & 2.39 & 2.70 & 2.31 & 2.60 & 1.78 & 3.26 \\
Wan2.2-I2V-A14B I2V & 3.36 & 1.95 & 4.00 & 2.33 & 2.74 & 2.50 & 3.65 & 1.56 & 3.39 \\
Seedance 2.5 I2V & 3.73 & 1.95 & 4.18 & 2.08 & 3.52 & 2.54 & 4.25 & 1.39 & 3.45 \\
LingBot-Video I2V & 5.00 & 2.24 & 3.27 & 3.00 & 3.00 & 2.25 & 3.40 & 1.65 & 3.48 \\
\toprule
\end{tabular}%
\end{table}

\begin{table}[!htbp]
\caption{Automatic evaluation mean score (1 to 5) of all 12 configurations on D5 (material properties and field states, 7 dims).}
\label{tab:alldim_d5}
\centering
\scriptsize
\begin{tabular}{lccccccc}
\toprule
\multicolumn{1}{c|}{\bf Config} & R05 & R06 & R07 & R13 & R14 & R18 & R25 \\
\hline
MiniMax H3 T2V & 2.56 & 3.62 & 2.27 & 1.48 & 1.76 & 2.41 & 1.52 \\
SANA-Video 2.0 T2V & 3.56 & 3.10 & 1.91 & 1.14 & 1.48 & 2.50 & 1.00 \\
Wan2.1-T2V-14B T2V & 1.83 & 1.95 & 1.32 & 1.57 & 1.10 & 2.82 & 1.00 \\
Wan2.2-T2V-A14B T2V & 1.33 & 1.90 & 1.59 & 1.86 & 1.29 & 2.91 & 1.17 \\
Seedance 2.5 T2V & 3.06 & 3.90 & 2.82 & 1.86 & 3.29 & 2.82 & 2.48 \\
LingBot-Video T2V & 2.56 & 2.71 & 1.68 & 1.86 & 1.86 & 2.73 & 1.70 \\
MiniMax H3 I2V & 2.44 & 3.76 & 2.32 & 1.67 & 2.33 & 2.86 & 1.17 \\
SANA-Video 2.0 I2V & 3.44 & 1.71 & 1.91 & 1.14 & 1.38 & 2.36 & 1.00 \\
Wan2.1-I2V-14B-720P I2V & 2.72 & 3.10 & 1.95 & 1.67 & 1.67 & 2.64 & 1.09 \\
Wan2.2-I2V-A14B I2V & 2.17 & 2.90 & 1.95 & 1.76 & 2.71 & 2.68 & 1.52 \\
Seedance 2.5 I2V & 3.00 & 3.81 & 2.77 & 1.76 & 2.90 & 2.68 & 2.24 \\
LingBot-Video I2V & 2.94 & 3.05 & 2.27 & 1.76 & 2.05 & 2.68 & 1.61 \\
\toprule
\end{tabular}%
\end{table}

\begin{table}[!htbp]
\caption{Automatic evaluation mean score (1 to 5) of all 12 configurations on D6 (kinematics, 5 dims).}
\label{tab:alldim_d6}
\centering
\scriptsize
\begin{tabular}{lccccc}
\toprule
\multicolumn{1}{c|}{\bf Config} & R15 & R16 & R17 & R20 & R21 \\
\hline
MiniMax H3 T2V & 4.22 & 3.10 & 1.19 & 3.45 & 3.33 \\
SANA-Video 2.0 T2V & 4.28 & 3.10 & 1.00 & 3.75 & 3.31 \\
Wan2.1-T2V-14B T2V & 4.33 & 3.05 & 1.05 & 4.00 & 3.62 \\
Wan2.2-T2V-A14B T2V & 4.44 & 2.95 & 1.05 & 3.95 & 3.18 \\
Seedance 2.5 T2V & 4.61 & 3.00 & 1.00 & 4.00 & 3.50 \\
LingBot-Video T2V & 4.22 & 2.95 & 1.00 & 3.85 & 3.20 \\
MiniMax H3 I2V & 4.33 & 2.90 & 1.00 & 3.85 & 3.06 \\
SANA-Video 2.0 I2V & 3.50 & 3.00 & 1.00 & 3.80 & 3.29 \\
Wan2.1-I2V-14B-720P I2V & 4.44 & 3.00 & 1.19 & 3.95 & 3.20 \\
Wan2.2-I2V-A14B I2V & 4.22 & 3.00 & 1.05 & 3.90 & 3.36 \\
Seedance 2.5 I2V & 4.61 & 3.00 & 1.05 & 4.00 & 3.23 \\
LingBot-Video I2V & 4.17 & 3.10 & 1.19 & 4.00 & 4.69 \\
\toprule
\end{tabular}%
\end{table}

\begin{table}[!htbp]
\caption{Automatic evaluation mean score (1 to 5) of all 12 configurations on D7 (mechanics and conservation, 4 dims).}
\label{tab:alldim_d7}
\centering
\scriptsize
\begin{tabular}{l|cccc}
\toprule
\multicolumn{1}{c|}{\bf Config} & R12 & R22 & R23 & R24 \\
\hline
MiniMax H3 T2V & 1.50 & 2.60 & 1.94 & 2.00 \\
SANA-Video 2.0 T2V & 1.36 & 2.80 & 1.83 & 1.00 \\
Wan2.1-T2V-14B T2V & 1.68 & 2.70 & 1.82 & 1.76 \\
Wan2.2-T2V-A14B T2V & 1.50 & 2.60 & 1.89 & 1.29 \\
Seedance 2.5 T2V & 1.36 & 2.70 & 2.83 & 1.90 \\
LingBot-Video T2V & 1.41 & 2.80 & 1.78 & 1.62 \\
MiniMax H3 I2V & 1.27 & 2.60 & 1.72 & 1.62 \\
SANA-Video 2.0 I2V & 1.10 & 2.65 & 1.94 & 1.43 \\
Wan2.1-I2V-14B-720P I2V & 1.48 & 2.75 & 1.94 & 1.67 \\
Wan2.2-I2V-A14B I2V & 1.59 & 2.70 & 1.82 & 1.67 \\
Seedance 2.5 I2V & 1.64 & 2.85 & 2.44 & 1.71 \\
LingBot-Video I2V & 1.23 & 2.75 & 1.89 & 2.43 \\
\toprule
\end{tabular}%
\end{table}

\subsection{Paired T2V/I2V Differences Across Models}
\label{sec:appendix:paired}

Table~\ref{tab:paired} reports the paired differences (I2V minus T2V) on all seven domains for the six model configurations.

\begin{table}[!htbp]
\caption{Paired differences (I2V minus T2V) on all seven domains across six model configurations.}
\label{tab:paired}
\begin{center}
\scriptsize
\setlength{\tabcolsep}{2pt}
\begin{tabular}{lccccccc}
\toprule
\bf Model & \bf CSF & \bf BC & \bf PTQ & \bf GS & \bf MFS & \bf KP & \bf MC \\
\hline
SANA-Video 2.0 & +0.9 & -16.1 & -9.8 & -11.2 & -6.1 & -4.2 & +0.8 \\
Wan2.1 & +19.8 & -2.6 & -4.2 & -9.7 & +11.6 & -1.4 & -0.8 \\
Wan2.2 & +15.5 & +1.5 & -1.5 & -3.4 & +13.0 & -0.3 & +3.1 \\
LingBot-Video & +7.4 & +2.0 & +0.1 & +8.0 & +4.5 & +9.6 & +4.3 \\
MiniMax H3 & +4.8 & +0.6 & +7.1 & -3.7 & +3.4 & -0.8 & -5.2 \\
Seedance 2.5 & +4.6 & -3.6 & -0.7 & -1.9 & -3.8 & -1.2 & -1.0 \\
\toprule
\end{tabular}
\end{center}
\end{table}

\section{Per-Dimension Scoring Principles}
\label{sec:appendix:scoring}

Each physical dimension is paired with a specialized vision foundation model to output numerical physical quantities, divided into three paradigms by evaluation method: \textbf{\textit{i)}} specialized-model quantitative evaluation; \textbf{\textit{ii)}} specialized-model measurement with MLLM judgment; \textbf{\textit{iii)}} pure MLLM semantic scoring. Table~\ref{tab:evalmap} gives the dimensions covered by each specialized model and the paradigm; the physical quantity, formula, and criterion extracted for each of the 25 physical dimensions are explained one by one in Section~\ref{sec:appendix:scoring:formulas}; the mapping from physical quantities to 1 to 5 threshold segments is in Section~\ref{sec:appendix:scoring:thresh} Table~\ref{tab:thresholds}. The D1 to D3 semantic/perceptual-layer dimensions are directly scored by MLLM or specialized perception models with 1 to 5 semantic anchors; the principle is in Section~\ref{sec:appendix:scoring:semantic}.

\subsection{Specialized Models and Dimension/Paradigm Mapping}
\label{sec:appendix:scoring:map}

Table~\ref{tab:evalmap} lists each evaluation model, the dimensions it covers, and the paradigm used. Geometric and kinematic dimensions are mainly output as numerical physical quantities by specialized models such as VGGT-$\Omega$~\citep{wang2026vggtomega}, StableNormal~\citep{ye2024stablenormal}, Depth Anything 3~\citep{lin2025depthanything3}, SAM 3~\citep{metasam2025sam3}, SpatialTracker V2~\citep{xiao2025spatialtrackerv2}, Any6D~\citep{lee2025any6d}, FlowSeek~\citep{poggi2025flowseek}, VFIMamba~\citep{zhang2024vfimamba}, DINOv3~\citep{metadino2025dino3}, and CLIP~\citep{radford2021clip} (paradigm i); dimensions such as light/shadow and object permanence are measured by ReasonX~\citep{dirik2025reasonx} / SAM 3 and then judged by MLLM (paradigm ii); material, thermal radiation, magnetic/electric, phase change, and all instruction-following/commonsense dimensions are directly semantically scored by Qwen3.8-27B~\citep{qwen2026qwen3.8} (paradigm iii).

\begin{table}[!htbp]
\caption{The dimensions covered by each specialized evaluation model and the evaluation paradigm. Paradigms: \textbf{\textit{i)}} specialized-model quantitative evaluation; \textbf{\textit{ii)}} specialized-model measurement + MLLM judgment; \textbf{\textit{iii)}} pure MLLM semantic scoring.}
\label{tab:evalmap}
\begin{center}
\small
\resizebox{\linewidth}{!}{%
\begin{tabular}{lll}
\toprule
\multicolumn{1}{c}{\bf MODEL} & \multicolumn{1}{c}{\bf COVERED DIMS} & \multicolumn{1}{c}{\bf PARADIGM} \\
\hline
VGGT-$\Omega$ & R01, R03, R08, R09, R11, R17, R18, R19 & (i) \\
Depth Anything 3 & R01, R04, R22, R23, R24 & (i) \\
StableNormal & R02, R13 & (i)/(ii) \\
SAM 3 & R01, R03, R04, R09, R11, R12, R15, R16, R21, R22, R23, R24 & (i)/(ii) \\
SpatialTracker V2 & R05, R12, R15, R16, R17, R20, R21, R22, R23, R24 & (i) \\
Any6D & R10, R17 & (i) \\
FlowSeek & R19 & (i) \\
VFIMamba & R20 & (i) \\
ReasonX & R13, R18 & (ii) \\
Qwen3.8-27B & R04, R06, R07, R10, R14, R25; IF-1 to IF-9, AC-1, CS-1 to CS-3 & (ii)/(iii) \\
DINOv3 & FQ-1, IF-10 & (i) \\
CLIP & TQ-4, IF-10 & (i) \\
FineVQ & FQ-2 & (i) \\
OpenCV/Farneback & TQ-1, TQ-2, TQ-3, TQ-5 & (i) \\
\toprule
\end{tabular}%
}
\end{center}
\end{table}

\subsection{Physical Quantities, Formulas, and Criteria}
\label{sec:appendix:scoring:formulas}

Below, the physical quantity, formula, and criterion extracted for each of the 25 physical dimensions are explained one by one by D4 to D7; the 1 to 5 threshold segments for each physical quantity are in Section~\ref{sec:appendix:scoring:thresh} Table~\ref{tab:thresholds}.

\paragraph{D4 Geometric and Spatial Consistency (L3 Physical Layer).}
\paragraph{R01 Depth.} We estimate the per-frame depth map using Depth Anything 3. Using the object occlusion regions segmented by SAM 3, we verify the depth-order accuracy, i.e., the depth of the occluded object should be greater than the depth of the occluder; the higher the depth-order accuracy, the higher the score.
\paragraph{R02 Surface Normal.} Estimated per-frame surface normal map by StableNormal, cross-validated with the geometric consistency provided by VGGT-$\Omega$ to check the consistency of normals with the depth gradient; the higher the consistency, the higher the score.
\paragraph{R03 Shape.} Per-frame 3D point clouds reconstructed by VGGT-$\Omega$; after Kabsch alignment\citep{kabsch1976solution} of cross-frame point clouds, the normalized Chamfer distance measures shape cross-frame self-consistency; non-standard geometry is semantically judged by Qwen3.8-27B. Higher self-consistency means more consistent shape.
\paragraph{R04 Absolute Scale.} Metric scale prior provided by Depth Anything 3, combined with the object size segmented by SAM 3, Qwen3.8-27B judges whether the object size falls in a reasonable range; the more reasonable, the higher the score.
\paragraph{R08 Multi-View Geometric Consistency.} For multiple views of the same scene (multiple frames produced by camera motion), VGGT-$\Omega$ reconstructs each separately, and the reprojection error and geometric self-consistency residual are computed to measure whether the 3D structure is consistent across views. Smaller residual is better.
\paragraph{R09 Relative Scale.} VGGT-$\Omega$ estimates the cross-frame bounding-box size of the object, and the coefficient of variation $\mathrm{CV}=\sigma/\mu$ of the size sequence is computed, with SAM 3 segmentation ensuring the same object corresponds across frames. Smaller CV means more stable relative scale.
\paragraph{R10 3D Pose.} Any6D estimates the six-degree-of-freedom pose of the object; combined with VGGT-$\Omega$ geometry, Qwen3.8-27B compares whether the final pose change matches the known pose change specified by the prompt; the higher the match, the higher the score.
\paragraph{R11 Spatial Relation.} VGGT-$\Omega$ back-projects to obtain the 3D coordinates of object centroids, computes the relative position and distance between objects, and compares consistency with the spatial relation described by the prompt (above/below/left/right/front/back/containment, etc.). Consistent gives a high score.
\paragraph{R19 Cross-Frame Geometric Consistency.} On the one hand, VGGT-$\Omega$ independently reconstructs point clouds for adjacent frames and computes the cross-frame Chamfer drift (relaxed for the VGGT-$\Omega$ reconstruction noise baseline); on the other hand, FlowSeek computes the coefficient of variation CV of the average endpoint error AEPE of per-frame optical flow. Both smaller indicate more stable cross-frame geometry.

\paragraph{D5 Material Properties and Field States (L3 Physical Layer).}
\paragraph{R05 Current Deformation State.} Per-frame point clouds reconstructed by VGGT-$\Omega$, combined with the object trajectory of SpatialTracker V2, after Kabsch alignment\citep{kabsch1976solution} compute the normalized Chamfer distance or rigid-body distance change rate to measure deformation; the threshold is higher than the VGGT-$\Omega$ reconstruction noise baseline (about 0.08 to 0.15) to avoid mistaking reconstruction noise for deformation. A smaller change rate means closer to rigid body.
\paragraph{R06 Material Properties.} Qwen3.8-27B performs a five-dimensional material check on the video (material category, hardness/softness, translucency, reflectivity, surface texture), judging item by item whether it conforms to physical commonsense. All dimensions reasonable gives a high score.
\paragraph{R07 Thermal Radiation.} Qwen3.8-27B performs a five-dimensional radiation check (heat source identification, heat conduction direction, heat-induced luminescence/color change, thermal convection/updraft, no abnormal radiation from low-temperature objects), judging item by item whether it conforms to thermodynamic laws. Conforming gives a high score.
\paragraph{R13 Light and Shadow Spatial Consistency.} ReasonX estimates the back-inferred light source direction of each object, combined with the normal of StableNormal, computes the variance of the back-inferred light source directions of each object in the same scene and the photometric/shadow-position residual (degrees); the smaller the variance and residual, the more self-consistent the light/shadow is spatially.
\paragraph{R14 Magnetic and Electric Interaction.} Qwen3.8-27B judges the correctness of electromagnetic/electrostatic phenomena in the video (like poles repel and unlike poles attract, the effect of the magnetic field on ferromagnetic objects, electrostatic adsorption of light small objects, the arc of the Van de Graaff generator along field lines, tip discharge, etc.). Correct phenomena give a high score.
\paragraph{R18 Light and Shadow Temporal Consistency.} ReasonX estimates the main light direction $L(t)$, and after camera-motion compensation to the world coordinate system, checks its temporal stability (degrees); the more stable the direction, the more consistent the light/shadow is temporally.
\paragraph{R25 Phase Transition.} Qwen3.8-27B judges the correctness of the phase transition process (conditions and direction of solid, liquid, and gas transition, temperature change due to latent heat, volume/morphology changes accompanying phase transition, etc.). A reasonable process gives a high score.

\paragraph{D6 Kinematics and Spatiotemporal Persistence (L3 Physical Layer).}
\paragraph{R15 Displacement.} SpatialTracker V2 (injecting query points with SAM 3 masks to obtain the object centroid trajectory) computes the L2 norm $\|x(t)\|_2$ (meters) of the object worldline, and SAM 3 detects whether the object suddenly appears/disappears and the symmetry of collision rebound. A trajectory consistent with physical expectation gives a high score.
\paragraph{R16 Velocity \& Acceleration.} The object centroid trajectory of SpatialTracker V2 computes velocity and the second derivative for acceleration, compared with the physical expectation for velocity error $\|\mathbf v-\hat{\mathbf v}\|$ (m/s), with 24 fps as the time base. Smaller error is better.
\paragraph{R17 Angular Velocity.} Any6D estimates the object rotation, combined with VGGT-$\Omega$ geometry to compute the angular velocity error $\|\omega-\hat\omega\|$ (rad/s), and additionally computes the coefficient of variation CV of angular velocity under free rotation (free rotation should be constant). Smaller error and CV are better.
\paragraph{R20 Motion Continuity.} VFIMamba and SpatialTracker V2 reconstruct odd frames by interpolating even frames and compare with the real odd frames to obtain the interpolation error, and simultaneously compute the jerk (rate of change of acceleration) of the 3D trajectory; smaller error and jerk mean more continuous motion.
\paragraph{R21 Object Permanence.} SAM 3 segmentation and VGGT-$\Omega$ geometry, combined with Qwen3.8-27B semantics, check whether object identity is maintained in the ``move away then move back'' scene, whether the depth front-back order constitutes a legal occlusion, and count the number of tracking ID jumps. Stable ID and no illegal occlusion give a high score.

\paragraph{D7 Mechanical Plausibility and Conservation Laws (L3 Physical Layer).}
\paragraph{R12 Stability.} VGGT-$\Omega$ gives the depth point cloud, RANSAC\citep{fischler1981random} fits the support plane, and SAM 3 locates the bottom contact point of the object and checks whether the center-of-mass projection of the object falls within the support polygon; inside the polygon means stable, otherwise it may tip over.
\paragraph{R22 Dynamic Effects of Forces.} Prompt-specific, scale-invariant checks cover Newton's first law, Galilean relativity, projectile motion, support contact, inclined-plane friction, sliding deceleration, bounce restitution, circular motion, and rolling. Metric scale from Depth Anything 3 is used to check acceleration against gravity. The residuals are aggregated into a score; smaller residuals are better.
\paragraph{R23 Momentum Conservation.} SpatialTracker V2 and SAM 3 obtain the object trajectory before and after collision. The mass ratio is directly specified in the prompt, and we compute the momentum residual $e=\|\mathbf p-\hat{\mathbf p}\|/\|\mathbf p\|$ accordingly. Smaller residual is better.
\paragraph{R24 Energy Conservation.} SpatialTracker V2, SAM 3, and the metric scale of Depth Anything 3 anchor the normalized coordinates to meters. The masses are obtained from the mass ratio specified in the prompt, so that kinetic energy $\mathrm{KE}$ and potential energy $\mathrm{PE}=gh$ can be computed without ambiguity; conservative scenarios should be approximately constant, and friction-involving scenarios should decrease monotonically and not rise spontaneously. The less deviation from constant/monotonic, the better.

\subsection{Mapping Physical Quantities to 1 to 5 Scores}
\label{sec:appendix:scoring:thresh}

Table~\ref{tab:thresholds} gives the 1 to 5 threshold segments for each physical quantity. The mapping rule is unified: for quantities where ``smaller is better'' (direction $\downarrow$), values successively below the 1st/2nd/3rd/4th segment threshold get 5/4/3/2 points, otherwise 1 point; for quantities where ``larger is better'' (direction $\uparrow$), values successively above the corresponding threshold get 5/4/3/2 points. Without simulation GT (T2V/I2V setting), geometric dimensions are mapped by the above rule using the physical quantities output by specialized models (e.g., R01 depth-order accuracy, R02 normal-depth consistency, R03 normalized Chamfer), while semantic/scale/pose dimensions (R04/R06/R07/R10/R14/R25, etc.) are directly given 1 to 5 by Qwen3.8-27B; the thresholds also connect to the rationality conventions of geometric evaluation (e.g., R02 normal-depth consistency takes $15^\circ$ as good), and relax the reconstruction noise baseline of specialized models (e.g., R05/R19).

\begin{table}[!htbp]
\caption{Mapping of physical quantities to threshold segments and 1 to 5 scores. Thresholds for each segment are listed in the order ``5 points to 1 point''; DIR marks the direction ($\downarrow$ smaller is better, $\uparrow$ larger is better).}
\label{tab:thresholds}
\begin{center}
\small
\resizebox{\linewidth}{!}{%
\begin{tabular}{lllc}
\toprule
\multicolumn{1}{c}{\bf METRIC (dim)} & \multicolumn{1}{c}{\bf THRESHOLDS (5$\to$1)} & \multicolumn{1}{c}{\bf UNIT} & \multicolumn{1}{c}{\bf DIR} \\
\hline
R01\_depth\_order\_accuracy (R01) & 1.00 / 0.75 / 0.50 / 0.25 & N/A & $\uparrow$ \\
R02\_normal\_depth\_consistency (R02) & 15 / 25 / 35 / 50 & $^\circ$ & $\downarrow$ \\
R03\_chamfer\_norm (R03) & 0.02 / 0.05 / 0.10 / 0.20 & N/A & $\downarrow$ \\
R04\_vlm\_reasonability (R04) & N/A (Qwen3.8-27B judges reasonable size range, directly 1 to 5) & N/A & $\uparrow$ \\
R05\_deform (R05) & 0.15 / 0.25 / 0.40 / 0.60 & N/A & $\downarrow$ \\
R09\_cv (R09) & 0.05 / 0.10 / 0.20 / 0.40 & N/A & $\downarrow$ \\
R10\_vlm\_pose\_match (R10) & N/A (Qwen3.8-27B compares with the pose change specified by the prompt, directly 1 to 5) & N/A & $\uparrow$ \\
R15\_worldline\_l2 (R15) & 0.10 / 0.30 / 0.50 / 1.00 & m & $\downarrow$ \\
R16\_vel\_err (R16) & 0.50 / 1.00 / 2.00 / 5.00 & m/s & $\downarrow$ \\
R17\_angvel\_err (R17) & 0.10 / 0.30 / 0.50 / 1.00 & rad/s & $\downarrow$ \\
R17\_rot\_cv (R17) & 0.05 / 0.10 / 0.20 / 0.50 & N/A & $\downarrow$ \\
R2x\_light\_angle (R13/R18) & 5 / 10 / 20 / 45 & $^\circ$ & $\downarrow$ \\
R19\_drift (R19) & 0.10 / 0.25 / 0.45 / 0.70 & N/A & $\downarrow$ \\
R19\_flow\_cv (R19) & 0.10 / 0.20 / 0.35 / 0.55 & N/A & $\downarrow$ \\
R22\_traj\_err (R22) & 0.10 / 0.30 / 0.50 / 1.00 & N/A & $\downarrow$ \\
R23\_mom\_err (R23) & 0.30 / 0.60 / 1.00 / 2.00 & N/A & $\downarrow$ \\
R24\_energy\_err (R24) & 0.50 / 1.00 / 2.00 / 5.00 & N/A & $\downarrow$ \\
FQ1\_subject\_cons (FQ-1) & 0.9964 / 0.9940 / 0.9910 / 0.9881 & N/A & $\uparrow$ \\
TQ1\_smoothness (TQ-1) & 0.98 / 0.95 / 0.90 / 0.80 & N/A & $\uparrow$ \\
TQ4\_bg\_cons (TQ-4) & 0.9912 / 0.9836 / 0.9741 / 0.9640 & N/A & $\uparrow$ \\
\toprule
\end{tabular}%
}
\end{center}
\end{table}

\subsection{Scoring Principles for Semantic and Perceptual Dimensions}
\label{sec:appendix:scoring:semantic}

D1 to D3 do not extract numerical physical quantities, but are directly scored by specialized perception models or MLLM with 1 to 5 semantic/numerical anchors, explained per dimension below.

\paragraph{D1 Condition and Semantic Fidelity (L1 Semantic Layer).}
\paragraph{IF-1 Subject Adherence.} Qwen3.8-27B judges whether the generated video contains the subject object specified by the prompt, assisted by CLIP/DINO objective reference scores; complete subject gives a high score.
\paragraph{IF-2 Attribute Adherence.} Qwen3.8-27B checks whether the attributes of the subject, such as color, material, and quantity, are consistent with the prompt (including color matching); if consistent, it gives a high score.
\paragraph{IF-4 Spatial Relation Adherence.} Qwen3.8-27B judges whether the spatial relations between objects (above/below/left/right/front/back/containment, etc.) conform to the prompt; conforming gives a high score.
\paragraph{IF-5 Style Adherence.} Qwen3.8-27B judges whether the artistic/appearance style of the frame conforms to the prompt; conforming gives a high score.
\paragraph{IF-7 Negation Adherence.} Qwen3.8-27B checks whether the negation constraints in the prompt (e.g., ``no X'', ``Y does not appear'') are obeyed; obeying gives a high score.
\paragraph{IF-8 Scene Matching.} Qwen3.8-27B judges whether the overall scene/environment is consistent with the scene described by the prompt; consistent gives a high score.
\paragraph{IF-9 Camera Motion Adherence.} Qwen3.8-27B judges whether the camera motion (push/pull/pan/move/orbit, etc.) conforms to the prompt; conforming gives a high score.
\paragraph{IF-10 Reference Condition Fidelity.} Evaluated only for I2V; aggregated from reference-image fidelity sub-indicators by DINOv3 + CLIP: IF-10a scene structure fidelity (DINOv3 feature similarity from reference image to first frame), IF-10c color/style continuity (DINOv3 Gram-matrix style distance, degraded to histogram KL divergence), IF-10d character face consistency (DINOv3 fine-feature tracking from reference image to video frame); higher fidelity gives a high score.

\paragraph{D2 Behavioral, Causal, and Commonsense Consistency (L1 Semantic Layer).}
\paragraph{IF-3 Action Adherence.} Qwen3.8-27B judges whether the action of the person/object conforms to the prompt (including human actions); conforming gives a high score.
\paragraph{IF-6 Temporal Adherence.} Qwen3.8-27B judges whether the event timing conforms to the prompt (including temporal semantics); conforming gives a high score.
\paragraph{AC-1 Action Causality.} Adopts counterfactual action pairs: generate two versions of the video, ``with action'' and ``without action'', for the same scene; SpatialTracker V2 tracks the 3D trajectory of the object, quantifies the causal effect (displacement amplification ratio, direction consistency), and then Qwen3.8-27B semantically verifies whether the result is reasonable; both effect and semantics reasonable give a high score.
\paragraph{CS-1 Physical Commonsense.} Qwen3.8-27B multiple-choice judgment plus explanatory reasonableness scoring, examining basic physical intuition (heavy objects fall, light objects float, fire rises, water flows down, shadow direction, etc.); reasonable gives a high score.
\paragraph{CS-2 Biological Commonsense.} Qwen3.8-27B judges the reasonableness of biological behavior (animal movement patterns, plant growth direction, human pose naturalness, etc.); reasonable gives a high score.
\paragraph{CS-3 Social Commonsense.} Qwen3.8-27B judges the reasonableness of scene logic (table and chair placement, naturalness of character interaction, daily causal consistency, etc.); reasonable gives a high score.

\paragraph{D3 Perceptual and Temporal Quality (L2 Perceptual Layer).}
\paragraph{FQ-1 Subject Consistency.} DINOv3 computes the cosine similarity of subject features between adjacent frames, mapped to 1 to 5 (threshold in Table~\ref{tab:thresholds}); higher similarity is better.
\paragraph{FQ-2 Aesthetic Quality.} FineVQ~\citep{duan2025finevq} (CVPR 2025, InternVL2-8B + regression head) scores the video aesthetic quality; the raw regression value is calibrated to 1 to 5 by distribution quantiles; higher score is better.
\paragraph{TQ-1 Motion Smoothness.} OpenCV/Farneback~\citep{bradski2000opencv,farneback2003two} optical flow computes motion acceleration smoothness $\max(0,1-\text{average acceleration}/2)$, mapped to 1 to 5 (threshold in Table~\ref{tab:thresholds}); smoother is better.
\paragraph{TQ-2 Temporal Flickering.} Measured by pixel differences in static regions of adjacent frames; smaller difference is better, mapped to 1 to 5.
\paragraph{TQ-3 Dynamic Degree.} Measured by the optical flow magnitude distribution; being consistent with the dynamic expectation of the prompt gives a high score.
\paragraph{TQ-4 Background Consistency.} CLIP computes the cosine similarity of background features between adjacent frames, mapped to 1 to 5 (threshold in Table~\ref{tab:thresholds}); higher similarity is better.
\paragraph{TQ-5 Imaging Quality.} Measured by Laplacian sharpness and exposure reasonableness; higher quality is better.

\section{Additional Prompt-Set Statistics}
\label{sec:appendix:promptstats}

The dataset contains 600 prompt groups, each paired with T2V and I2V versions. Table~\ref{tab:prompts} summarizes their composition and the share of embodied scenarios.

\begin{table}[!htbp]
\caption{Statistics of the 600 prompt groups: category counts, embodied samples, and modality pairing.}
\label{tab:prompts}
\begin{center}
\small
\begin{tabular}{lccc}
\toprule
\multicolumn{1}{c}{\bf Prompt Category} & \multicolumn{1}{c}{\bf Count} & \multicolumn{1}{c}{\bf Embodied Samples} & \multicolumn{1}{c}{\bf Description} \\
\hline
Dimension-specific & 230 & 0 & 5 per dimension \\
Core combinations & 230 & 40 & 23 paired dimensions \\
High-difficulty supplement & 140 & 140 & All involve embodied agents \\
Total & 600 & 180 (30\%) & Paired T2V/I2V versions \\
\toprule
\end{tabular}
\end{center}
\end{table}

\section{Generation Protocol Details}
\label{sec:appendix:protocol}

All models generate all 600 prompt groups one by one. T2V is conditioned on text, I2V is conditioned on the first-frame reference image synthesized by GPT-image-2, ensuring T2V/I2V share the same initial scene (see main text Section~\ref{sec:method:prompt}). The unified output specification is 720p, about 10 seconds (about 240 frames, 24 fps), with a fixed random seed; each model uses its officially recommended inference configuration and weights. Generated videos are uniformly resampled to the resolution and frame rate required by the evaluators before evaluation. The generation scripts, model versions, and hyperparameters will be released with the code to support third-party reproducibility.

\section{Evaluation Cost and Reproducibility}
\label{sec:appendix:repro}

For reproducibility, the versions and weights of specialized models and the random seed are fixed during evaluation. The experiments cover 12 model configurations and 600 prompt groups (about 7,200 videos); the evaluation models run on a single H20 GPU (95 GB). Per-video runtime and total GPU hours vary with hardware and the evaluated dimensions. Further statistics, the runtime environment, and model-weight versions will be provided with the code release.

\end{document}